\PassOptionsToPackage{table}{xcolor}
\documentclass[10pt,twocolumn,letterpaper]{article}
\usepackage[pagenumbers]{iccv} 

\newcommand{\ours}{HRScene}

\usepackage{pifont}
\newcommand{\cmark}{\ding{51}}
\newcommand{\xmark}{\ding{55}}

\usepackage{booktabs}
\usepackage[normalem]{ulem}
\useunder{\uline}{\ul}{}

\usepackage{CJKutf8}

\usepackage{lscape}
\usepackage{amsmath}

\definecolor{iccvblue}{rgb}{0.21,0.49,0.74}
\usepackage[pagebackref,breaklinks,colorlinks,allcolors=iccvblue]{hyperref}

\def\confName{ICCV}
\def\confYear{2025}

\title{ \ours: How Far Are VLMs from Effective High-Resolution Image Understanding?}

\author{
Yusen Zhang, Wenliang Zheng, Aashrith Madasu, Peng Shi, Ryo Kamoi, Hao Zhou, \\ Zhuoyang Zou, Shu Zhao, Sarkar Snigdha Sarathi Das,
Vipul Gupta, Xiaoxin Lu, Nan Zhang, \\ Ranran Haoran Zhang, Avitej Iyer, Renze Lou, Wenpeng Yin, Rui Zhang \\
Penn State University\\
{\tt\small \{yfz5488, rmz5227\}@psu.edu}
\\[2mm]
\textbf{\url{https://yszh8.github.io/hrscene/}}
}

\begin{document}
\maketitle
\begin{abstract}
High-resolution image (HRI) understanding aims to process images with a large number of pixels, such as pathological images and agricultural aerial images, both of which can exceed 1 million pixels. Vision Large Language Models (VLMs) can allegedly handle HRIs, however, there is a lack of a comprehensive benchmark for VLMs to evaluate HRI understanding. To address this gap, we introduce HRScene, a novel unified benchmark for HRI understanding with rich scenes. HRScene incorporates 25 real-world datasets and 2 synthetic diagnostic datasets with resolutions ranging from 1,024 $\times$ 1,024 to 35,503 $\times$ 26,627. HRScene is collected and re-annotated by 10 graduate-level annotators, covering 25 scenarios, ranging from microscopic to radiology images, street views, long-range pictures, and telescope images. It includes HRIs of real-world objects, scanned documents, and composite multi-image. The two diagnostic evaluation datasets are synthesized by combining the target image with the gold answer and distracting images in different orders, assessing how well models utilize regions in HRI. We conduct extensive experiments involving 28 VLMs, including Gemini 2.0 Flash and GPT-4o. Experiments on HRScene show that current VLMs achieve an average accuracy of around 50\% on real-world tasks, revealing significant gaps in HRI understanding. Results on synthetic datasets reveal that VLMs struggle to effectively utilize HRI regions, showing significant Regional Divergence and lost-in-middle, shedding light on future research.


\end{abstract}
    
\section{Introduction}
\label{sec:intro}
\begin{figure*}[t]
    \centering
    \begin{subfigure}{0.31\textwidth}
        \centering
        \includegraphics[width=\linewidth]{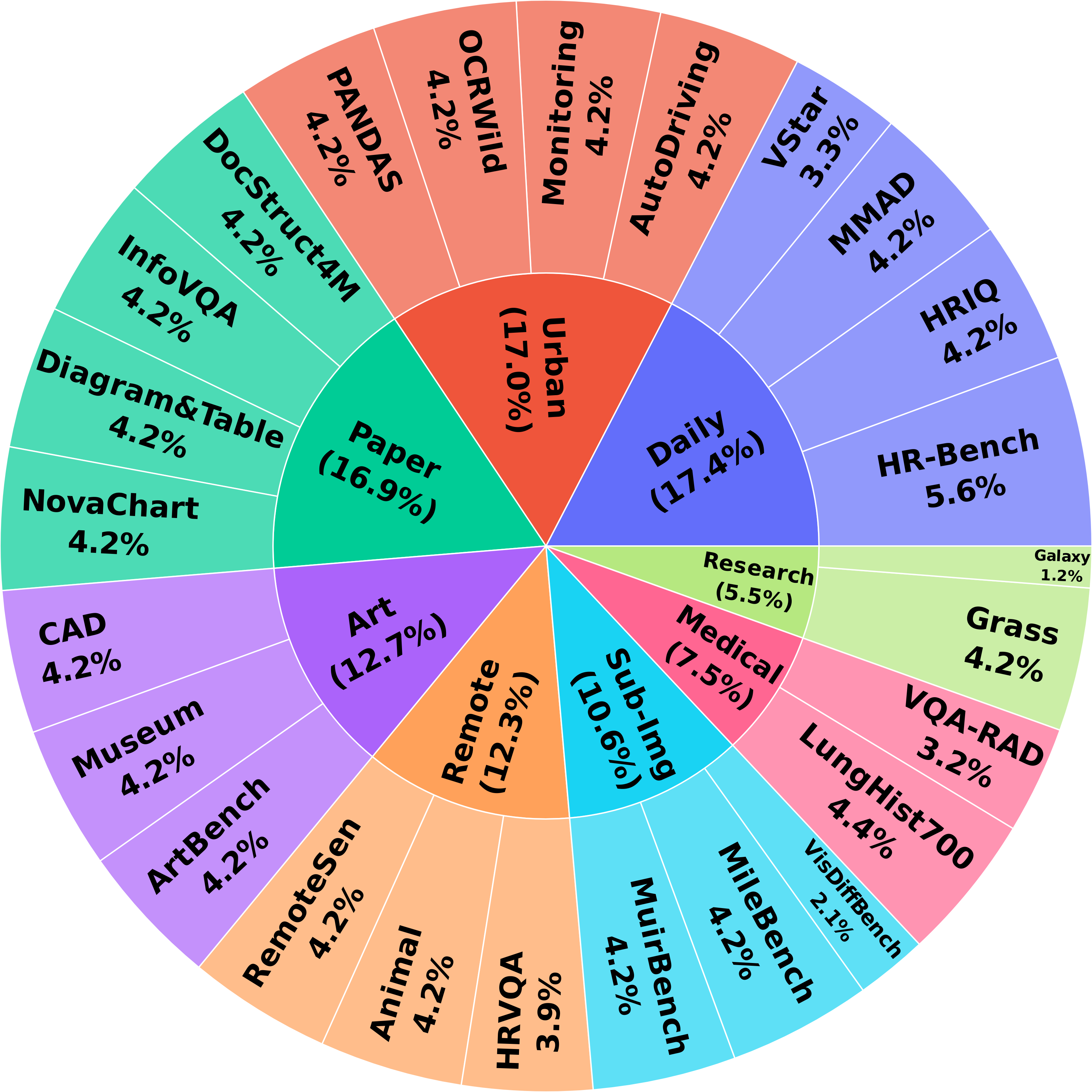}
        \caption{}
        \label{fig:figure1a}
    \end{subfigure}
    \hfill
    \begin{subfigure}{0.32\textwidth}
        \centering
        \includegraphics[width=\linewidth]{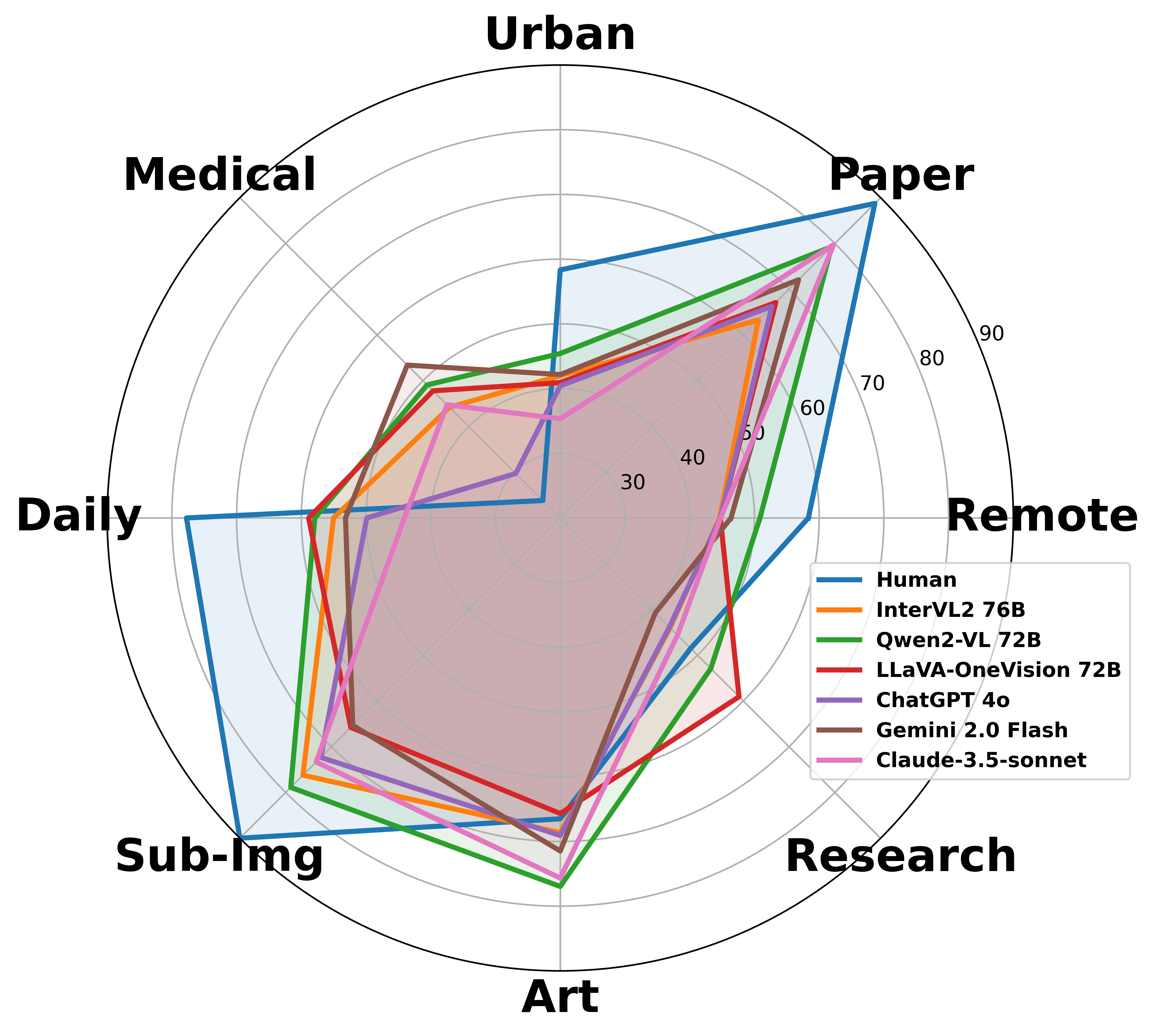}
        \caption{}
        \label{fig:figure1b}
    \end{subfigure}
    \hfill
    \begin{subfigure}{0.33\textwidth}
        \centering
        \includegraphics[width=\linewidth]{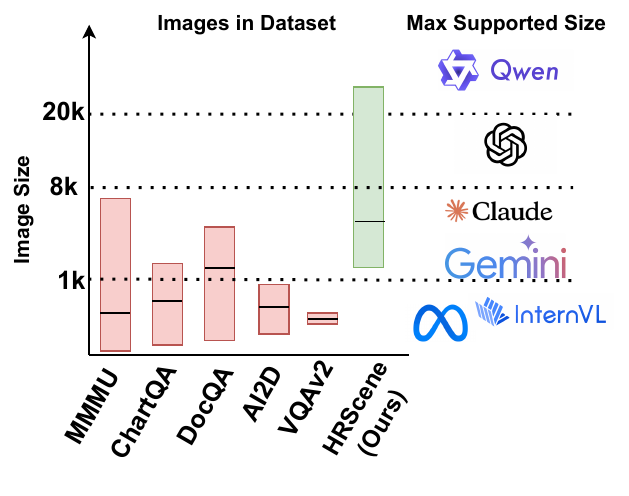}
        \caption{}
        \label{fig:figure1c}
    \end{subfigure}
    \caption{(a) Overview taxonomy of the \ours. (b) Performance of some VLMs on \ours. (c) Comparison between the benchmarks that the mainstream VLMs are evaluated on and \ours. The y-axis is the $\sqrt{\text{total pixel}}$. The boxes/icons indicate the image resolution they contain/support. The black lines inside each box show the average resolutions.}
    \label{fig:figure1}
\end{figure*}

High-resolution image (HRI) understanding aims to process images with a large number of pixels~\cite{bakhtiarnia2024efficient}. It plays an important role in numerous scenarios, such as pathology~\cite{Diosdado2024-zg}, autonomous driving~\cite{zhang2024mme}, and large document understanding~\cite{hu2024mplug,mathew2022infographicvqa,hu2024novachart}. With the development of Vision Large Language Models (VLMs), automatic processing of HRIs has been a promising direction~\cite{wang2024qwen2}. As shown in Figure~\ref{fig:figure1c}, Gemini~\cite{gemini}, Claude~\cite{anthropic2024claude}, and GPT~\cite{achiam2023gpt} can support images exceeding 1k resolution, enabling a wide range of real-world applications, such as 24/7 street monitoring~\cite{zhang2024mme}, galaxy research~\cite{imam2024cosmoclip}, and radiology analysis~\cite{lau2018dataset}.

However, even though existing VLMs can allegedly handle inputs of high-resolution images, there is a lack of comprehensive HRI benchmark, hindering objective calibration and measurement of progress on the effectiveness of HRI understanding. First, HRI evaluation is often missing from the official reports of mainstream VLMs.
Figure~\ref{fig:figure1c} lists most of the vision-based benchmarks that VLMs are evaluated on, such as MMMU~\cite{yue2024mmmu}, VQAv2~\cite{goyal2017making}, and AI2D~\cite{kembhavi2016diagram}. Their average resolution is typically below 1k, making them unsuitable for HRI evaluation. Moreover, there is no comprehensive real-world or diagnostic benchmark for HRIs. As shown in Table~\ref{tab:compare_work}, the existing real-world datasets with HRIs tend to either focus on specific scenarios, like long-range images~\cite{hrbench}, or particular resolutions, such as 8k~\cite{zhang2024mme}. The current diagnostic evaluation, namely, Multi-modal Neelde-in-the-Haystack (NIAH), primarily focuses on long text~\cite{liu-etal-2024-lost} or a mixture of low-resolution images~\cite{wang2025needle}.

To address this gap, we introduce \ours, a unified benchmark for HRI understanding, covering diverse real-world scenes. \ours~incorporates 25 real-world tasks with resolutions ranging from 1024 $\times$ 1024 to 35,503 $\times$ 26,627, and 2 synthetic diagnostic datasets with 1k to 4k resolution. As shown in Figure~\ref{fig:figure1a}, we propose a task taxonomy to guide the development of \ours~: (1) we identify 8 categories of HRI tasks:  Daily pictures, Urban planning, Paper scanned images, Artworks, Multi-subimages, Remote sensing, Medical Diagnosing, and Research understanding. (2) We focus on the 25 real-world scenes distributed across these categories, such as street monitoring and medical image understanding. (3) Each scene evaluates diverse capabilities of LLMs, such as counting, temporal and semantic reasoning, holistic image judgment, visual retrieval, spatial relations, and small object detection.
\begin{table}[!ht]
\caption{Comparison with existing real-world benchmarks and Multi-modal NIAH diagnosis.}
\label{tab:my-table}
\resizebox{\linewidth}{!}{
\begin{tabular}{@{}lrrrr@{}}
\toprule
\textbf{Benchmark} & Scenes & Easy Eval & Highest Res & Avg Res \\\midrule
MME~\cite{zhang2024mme} & \multicolumn{1}{r}{5} & \cmark & 5304 $\times$ 7952 & 2000 $\times$ 1500 \\
HR-Bench~\cite{hrbench} & \multicolumn{1}{r}{1} & \xmark & 8000 $\times$ 8000 & 8000 $\times$ 8000 \\
\ours~(ours) & \multicolumn{1}{r}{25} & \cmark & 35503 $\times$ 26627 & 4828 $\times$ 4078 \\ 
\midrule\midrule
\textbf{NIAH} & High-Res& Multi-Res & Real Img & Needle in \\ \midrule 
MM-NIAH~\cite{wang2025needle} & \xmark & \xmark & \cmark &  text \\
MileBench~\cite{song2024milebench} & \xmark & \xmark & \cmark & text \\
Visual Haystack~\cite{wu2024visual} & \xmark & \xmark & \xmark & image \\
\ours~(ours) & \cmark & \cmark & \cmark & image \\\bottomrule
\end{tabular}}
\label{tab:compare_work}
\end{table}
\ours~is collected from 25 existing data resources, and 8 of them are re-annotated by 10 graduate-level annotators, with diverse view scales, ranging from microscope to radiology, street views, long-range, and telescope images. It contains high-resolution images of real objects, electronic documents, and composite multi-subimages. Besides, six datasets require domain-expert knowledge, while the remaining 19 belong to general domains. The diagnostic dataset is synthesized by combining the target image with the gold answer and visually similar distractors arranged in different orders to assess HRI utilization. Overall, \ours~comprises 7,068 images, with 2,008 of them being re-annotated.

We conduct extensive experiments to evaluate the HRI understanding of 28 popular VLMs, including 6 proprietary VLMs: GPT~\cite{gpt4o}, Claude~\cite{anthropic2024claude}, and Gemini~\cite{team2024gemini} families, and 22 open-sourced models, including InternVL2~\cite{chen2024internvl}, DeepSeek~\cite{deepseekvl2}, Phi~\cite{abdin2024phi}, Qwen~\cite{Qwen2VL}, MolMo~\cite{deitke2024molmo}, LLava~\cite{llava}, and Llama~\cite{llama} families. As shown in Figure~\ref{fig:figure1b}, experiments on real-world tasks demonstrate that current VLMs perform modestly, with an average accuracy of around 50\%, highlighting substantial challenges of \ours. Besides, we also provide the human performance of all real-world datasets by engaging graduate-level annotators to annotate 750 image-question pairs. Our synthetic datasets provide a fine-grained understanding of VLM performance, revealing two robust issues across VLMs, including regional divergence and lost-in-the-middle, shedding light on future improvement directions. 

Our contributions are: (1) we propose \ours, a unified benchmark with 25 real-world and 2 diagnostic datasets; (2) we benchmark 28 models on \ours~ and show the significant performance gap; (3) we discover two salient issues of VLMs, including regional divergence and the lost-in-the-middle.


\begin{figure}
    \centering
    \includegraphics[width=\linewidth]{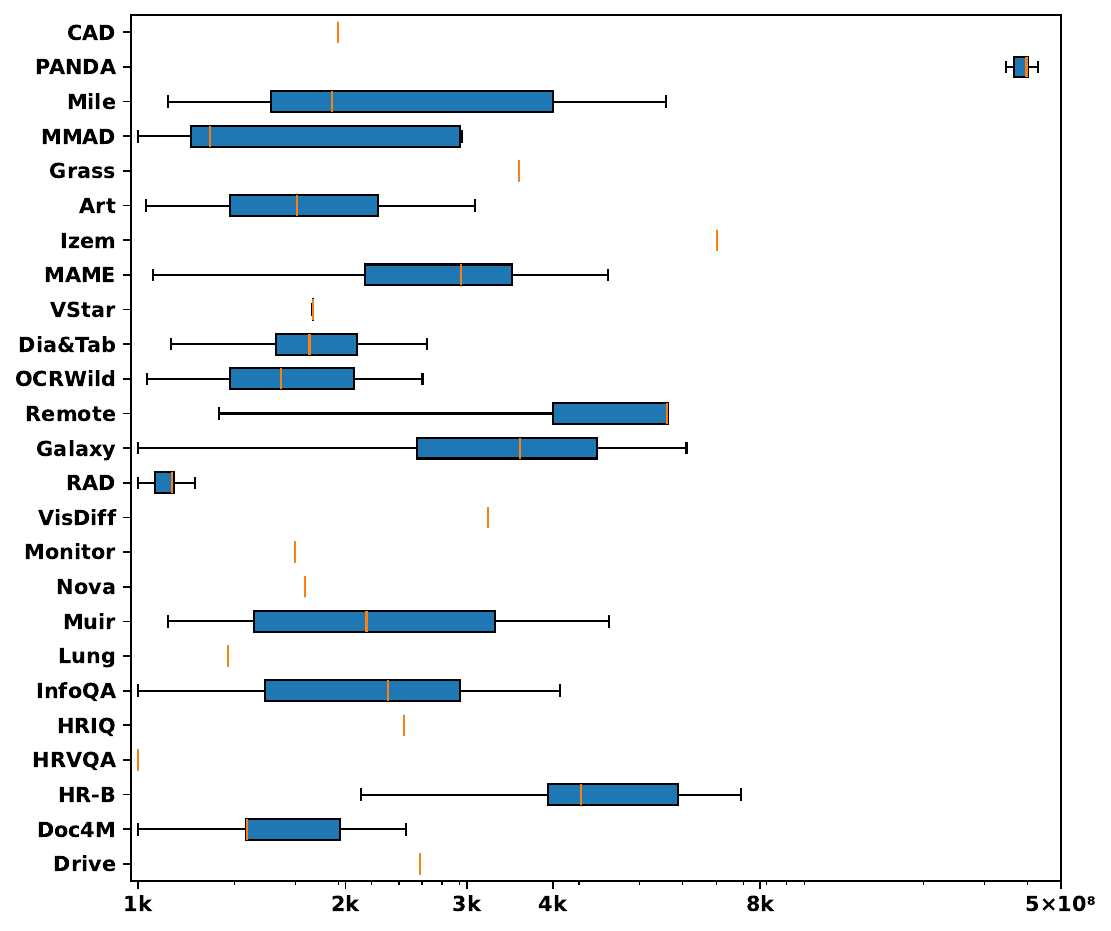}
    \caption{Distribution of resolution of each dataset. X-axis is the resolution and $n$k indicates the resolution is at least $n^2*10^6$ pixels.}
    \label{fig:res_distribution}
\end{figure}
\section{Related Work}
\noindent\textbf{VLMs for High-resolution Image.} Recent advances in vision-language models have demonstrated remarkable capabilities in understanding and reasoning about visual content~\cite{gpt4o,gemini,llama,llava}. However, processing high-resolution images remains a significant challenge due to computational constraints and the need to capture both fine-grained details and global context~\cite{hrsuvery}. Two main categories of approaches have emerged to address these challenges. The first category employs a dual encoder architecture that processes the high-resolution and low-resolution of the same image in parallel~\cite{cogagent,deepseekvl,mra}. A low-resolution encoder, typically CLIP~\cite{clip}, captures coarse-grained features for global understanding, while a high-resolution encoder based on convolutional neural networks~\cite{convnext} or object segmentation models~\cite{sam} preserves fine-grained details. The resulting high-resolution and low-resolution tokens are then fused or concatenated before being passed to large language models. This design achieves computational efficiency while leveraging both global and local visual features. The second category utilizes a splitting strategy~\cite{minimonkey,monkey,mm1,llavauhd,internlmx4khd,mplugdocowl15,deepseekvl2}. A high-resolution image is first downsampled and processed by a vision encoder to capture the global context, while the original image is then divided into multiple sub-images processed by a vision encoder with the same resolution to extract local features. It enables efficient processing of high-resolution images while preserving fine details, proving particularly effective for tasks requiring detailed local analysis and global scene understanding. In this work, we propose \ours, a benchmark with diverse scenes that challenge both global and local understanding capabilities of VLMs. 

\noindent\textbf{Evaluating High-resolution Image Understanding.}
High-resolution image understanding has been an important topic in computer vision~\cite{bakhtiarnia2024efficient}. Tradition research has usually focused on downstream tasks such as classification and detection, such as Crowd Counting~\cite{idrees2013multi,7780439,idrees2018composition,wang2020panda,sindagi2020jhu}, Autonomous Driving~\cite{cordts2016cityscapes,ros2016synthia,huang2018apolloscape,li2020towards,yu2020bdd100k}, Aerial Image Classification~\cite{christie2018functional}, and Pathology~\cite{veta2019predicting,aresta2019bach,cancer2012comprehensive,yang2020deep}. With the development of VLMs, more benchmarks have been introduced based on the stronger capability of VLMs, these tasks usually incorporate text instructions, logical reasoning, and complex question answering. For instance, MME-Realworld~\cite{zhang2024mme} proposes five tasks with small objects for VLMs to recognize and answer the given question; MileBench focuses on the ability of VLMs to discern relationships of multiple images that are possibly temporal or semantically interconnected; and MuirBench contains 12 diverse multi-image tasks (e.g., scene understanding, ordering) with 10 multi-image relations. However, these works are separately proposed and focus on limited scenes, failing to evaluate the VLMs' capability to HRI understanding comprehensively.

\noindent\textbf{Needle-in-the-Haystack.} Needle in the Haystack (NIAH) test is initially proposed for long text input~\cite{liu-etal-2024-lost}. It is used to evaluate the LLM capability in long context understanding and mark the text regions that LLMs fail to understand. Recently, this NIAH test has been extended for Multi-modality settings. MM-NIAH~\cite{wang2025needle} evaluates VLMs with a question on the synthetic sub-image embedded in a larger whole image. MileBench~\cite{song2024milebench} mixes the text and image input and asks the model to retrieve both text and image information. Visual Haystack~\cite{wu2024visual} feeds VLM with multiple images and tests the model's robustness to the permutation of this input. Different from these works, we analyze the HRI and permute the sub-images inside one large image. 

\begin{table}[!ht]
\caption{General Statistics of \ours.}

\label{tab:statis}
\resizebox{\linewidth}{!}{
\begin{tabular}{@{}lr|lr|lr@{}}
\toprule
Samples & \multicolumn{1}{l|}{\#} & Tasks & \multicolumn{1}{l|}{\#} & Elements & \multicolumn{1}{l}{\#} \\ \midrule
Total & 7068 & Total & 27 & Images & 7068 \\
Reannotated & 2005 & Real-world & 25 & Questions & 5807 \\
Scratch & 384 & Synthetic & 2 & Options & 34372 \\ \bottomrule
\end{tabular}}
\end{table}
\begin{figure*}
    \centering
    \includegraphics[width=\linewidth]{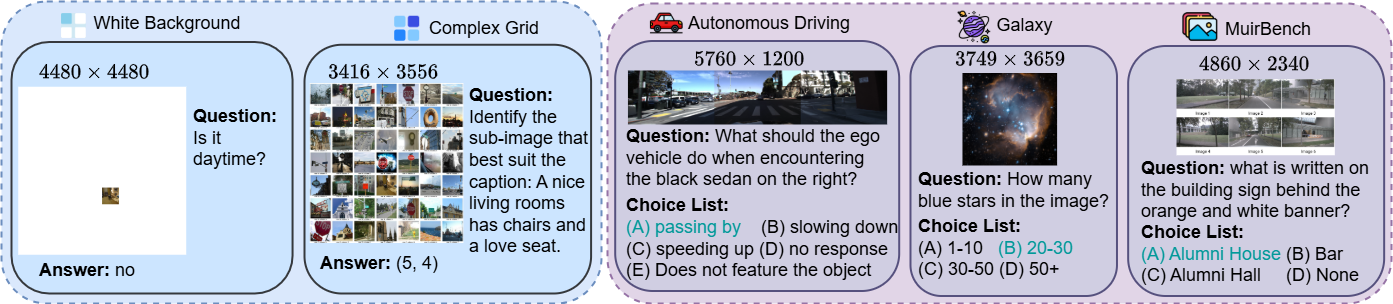}
    \caption{Some examples of \ours. Blue ones are diagnostic datasets and purple ones are real-world datasets. }
    \label{fig:example}
\end{figure*}

\section{The \ours~Benchmark}
This section outlines our benchmark construction process, including a brief overview of each adopted category and the manual efforts to ensure their adoption and maintain high data quality.
\subsection{Collection Guidelines}
As mentioned previously, there is a significant gap in the lack of unified, comprehensive, and easy-to-use HRI benchmarks for VLMs. \ours~ is motivated to address this gap, offering a high-quality evaluation benchmark for HRI understanding, and pushing VLMs to a general-purpose HRI processor. 
To create a high-quality benchmark, we consider the following guidelines for the creation: 
(1) consider possible real-world scenes where users need VLMs to process HRIs. Think in broader categories and taxonomy.
(2) Create comprehensive and most important tasks for each category without duplication of tested capability. 
(3) Ensure \ours~is easy to use. Each task does not need too many data points while being easy to verify the correctness. 

The taxonomy for this work is introduced as follows: First, we identify 8 categories of scenes: Daily pictures, Urban planning, Paper scanned images, Artwork, Multi-sub-images, Remote sensing, Medical Diagnosing, and Research understanding. 
Second, we cover a wide array of camera types, from microscopes to daily cameras, long-range cameras, x-ray devices, remote sensing cameras, and telescopes. 
Last but not least, we enrich the format and requirements of the datasets by including both single-image and multi-image data, expert and non-expert level QA, as well as small object detection and global image understanding.

\subsection{Data Collection}
In general, for all tasks, we ensure the selection of a balanced subset by uniformly sampling across all categories, such as question types, subtasks, and domains. We also filter out any images that do not meet the minimum resolution threshold of 1024x1024 (1K). After sampling, all datasets undergo human inspection, and any low-quality data points are removed. The supplementary materials contain more details about all datasets.
\textbf{Daily pictures}
include the pictures captured by daily users with high-resolution cameras, including \textit{HRIQ} \cite{huang2024high}, designed for Blind Image Quality Assessment, \textit{VStar} \cite{wu2024v}, designed for image search of objects in daily images, \textit{MMAD} \cite{jiang2024mmad}, a dataset for abnormal detection of daily items, and \textit{HR-Bench} \cite{wang2024divide}, for long-range picture question answering. 
\textbf{Urban planning} contains the scenes that are helpful for intelligent urban construction: \textit{Autonomous Driving} \cite{zhang2024mme} contains images from car camera, \textit{Monitoring} \cite{zhang2024mme} contains street camera images, \textit{OCRWild} \cite{zhang2024mme} contains OCR of street signs, and \textit{PANDA} \cite{wang2020panda} is the monitor images from a crowded environment.
\textbf{Paper scanned images} consist of scanned images from paper or documents, including structural documents (\textit{DocStruct4M} \cite{hu2024mplug}), rich graphic documents (\textit{InfoVQA} \cite{mathew2022infographicvqa}), chart data (\textit{NovaChart} \cite{hu2024novachart}), and complex diagrams and tables (\textit{Diagram\&Table} \cite{zhang2024mme}).
\textbf{Artwork} tests the model's capability to understand art or design. \textit{MAME} \cite{pares2022mame} contains the artwork images from museums, \textit{ArtBench} \cite{liao2022artbench} is a dataset with various paintings of different themes. \textit{CAD} \cite{fan2021floorplancad} is a floor plan dataset for interior design.
\textbf{Multi-sub-images} contain images composed of multiple small sub-images. This includes \textit{VisDiffBench} \cite{dunlap2024describing} that requires telling the difference between two groups of images. \textit{MileBench} \cite{song2024milebench} contains frames from a video to form a larger image. \textit{MuirBench} \cite{wang2024muirbench} focuses on the multi-image of various scenes and tests diverse capabilities, such as spatial relations.
\textbf{Remote sensing} is geographic images captured by remote devices. This category includes \textit{RemoteSen} \cite{zhang2024mme} for aerial image QA and \textit{HRVQA} \cite{li2024hrvqa} for remote small object detection. \textit{Animal} \cite{Izembek2022} contains aerial images of geese from Izembek Lagoon in Alaska.
\textbf{Medical Diagnosing} is for medical purposes. \textit{VQA-RAD} \cite{Lau2018} is a VQA dataset for radiology images of various types (X-rays and CT scans). \textit{LungHist700} \cite{LungHist700} is a collection of histopathological lung tissue images.
\textbf{Research} includes the images for expert-domain research. \textit{Grass} \cite{Grass_Urochloa} is phenological stage classification and raceme counting of Urochloa images. \textit{Galaxy}\footnote{\url{https://esahubble.org/images/}} contains the images from Hubble telescopes.

\subsection{Data Reannotation}
\noindent\textbf{Task annotation} Although most of the datasets contain annotations that can be directly used, some of them are not easy to evaluate or do not have labels. To this end, we ask 10 graduate-level annotators to reannotate 8 datasets. For six of them, we construct several wrong options so that each sample has at least 4 options in total. The wrong options are designed to be distracting and valid. Also, for two of the datasets, we annotate from scratch, with one question-answer pair for each image. For datasets with numeric answers, we automatically generate random numbers $r \in [-a,+a]$ multiple times to ensure 4 options, where $a$ is the correct answer.

\begin{table*}[ht!]
\centering
\caption{Overall performance of all models on real-world datasets of \ours. The models are grouped according to the parameter sizes. \textbf{Bold} indicates global best performance, while {\ul underline} represents the best of the group. Avg. is the mean value of the column/row.  Each category represents the average score of every sample of the corresponding category in the \textit{test} set.}
\label{tab:overall}
\resizebox{0.88\linewidth}{!}{

\begin{tabular}{@{}ccccccccccc@{}}
\toprule
Model & Avg. \textit{testmini} & Art & Daily & Medical & Paper & Remote & Research & Sub-Img & Urban & Avg. \\ \midrule
Random & -- & 15.32 & 26.65 & 22.33 & 23.44 & 22.38 & 21.12 & 25.42 & 21.25 & 22.46 \\ \midrule
Phi3.5 4B & 44.78 & 57.32 & 52.36 & 32.66 & 55.39 & 37.87 & 40.84 & 57.75 & 35.74 & 47.35 \\
DeepSeek-Janus 7B & 38.65 & 53.78 & 41.44 & 35.88 & 36.00 & 35.82 & 37.36 & 47.17 & 34.48 & 40.19 \\
MolMo 7B-D & 44.43 & 47.89 & 54.68 & 31.16 & 52.47 & 42.73 & 31.70 & 52.97 & 38.16 & 45.95 \\
InternVL2 1B & 35.58 & 31.07 & 36.83 & 30.17 & 32.37 & 30.34 & 26.22 & 37.62 & 25.59 & 31.63 \\
InternVL2 2B & 39.43 & 49.60 & 45.34 & 32.58 & 45.72 & 39.60 & 27.58 & 42.53 & 36.27 & 41.47 \\
InternVL2 4B & 47.46 & 60.74 & 47.18 & {\ul 42.02} & 57.91 & 35.96 & 35.52 & 66.16 & {\ul 40.77} & 49.23 \\
InternVL2 8B & 49.61 & 61.32 & 53.84 & 40.45 & 58.20 & 37.48 & 40.67 & 64.87 & 38.16 & 50.29 \\
Qwen2-VL 7B & {\ul 55.39} & {\ul 69.46} & {\ul 64.20} & 40.40 & {\ul 64.62} & {\ul 50.60} & 36.69 & {\ul 71.42} & 40.17 & {\ul 56.65} \\
Llava-HR 7B & 30.86 & 39.58 & 41.18 & 27.36 & 31.47 & 35.56 & 27.06 & 36.62 & 30.64 & 34.56 \\
Llava-Next 8B & 40.15 & 47.44 & 47.54 & 32.84 & 39.84 & 46.91 & 28.41 & 54.74 & 33.31 & 42.34 \\
Llava-Next 13B & 40.06 & 43.11 & 46.02 & 36.53 & 46.28 & 34.31 & 30.40 & 54.35 & 33.74 & 41.47 \\ 
Llama3.2 11B & 49.64 & 65.62 & 49.51 & 41.85 & 59.45 & 43.99 & {\ul 42.65} & 59.16 & 38.19 & 50.71 \\\midrule
InternVL2 26B & 53.86 & 66.97 & 57.95 & {\ul 38.15} & 64.62 & 42.58 & 36.04 & 68.27 & {\ul \textbf{45.90}} & 54.69 \\
DeepSeek-VL2 27B & 53.09 & 71.83 & 58.78 & 35.86 & 61.84 & {\ul 46.16} & 34.88 & 66.05 & 44.59 & 54.73 \\
InternVL2 40B & {\ul 60.55} & {\ul 74.35} & {\ul 62.67} & 38.10 & {\ul 70.89} & 44.16 & 43.15 & {\ul 74.10} & 44.40 & {\ul 58.45} \\
Llava-Next 34B & 50.42 & 64.74 & 55.59 & 36.90 & 57.28 & 42.46 & {\ul 51.45} & 62.60 & 35.54 & 51.13 \\ \midrule
InternVL2 76B & 55.38 & 69.74 & 61.68 & 38.08 & 67.07 & 43.50 & 29.40 & 73.68 & 41.28 & 55.64 \\
Llama3.2 90B & 54.88 & 70.66 & 51.90 & 40.02 & 64.31 & 44.85 & 31.56 & 63.48 & 39.28 & 52.60 \\
Llava-Next 72B & 47.00 & 61.00 & 54.20 & 36.15 & 57.19 & 42.03 & 31.56 & 59.13 & 34.74 & 48.69 \\
Llava-OneVision 72B & 55.96 & 68.26 & 64.64 & 42.00 & 68.91 & 46.18 & 52.12 & 68.75 & 40.28 & 57.45 \\
MolMo 72B & 54.02 & 60.30 & 58.16 & 35.91 & 63.67 & 52.16 & {\ul \textbf{55.59}} & 64.97 & 43.99 & 55.12 \\
Qwen2-VL 72B & {\ul \textbf{62.03}} & {\ul 75.85} & {\ul \textbf{66.20}} & {\ul 43.69} & {\ul 78.13} & {\ul \textbf{52.48}} & 39.36 & {\ul \textbf{74.89}} & {\ul 44.66} & {\ul \textbf{61.85}} \\ \midrule
Calude3 Haiku & 40.84 & 57.90 & 37.14 & 27.66 & 55.08 & 30.05 & 29.24 & 57.43 & 27.67 & 41.34 \\
Calude3.5 Sonnect & 55.49 & 75.26 & 50.06 & 40.41 & {\ul \textbf{78.85}} & 40.63 & 26.57 & {\ul 69.70} & 34.29 & 54.37 \\
Gemini1.5 Pro & 52.00 & 73.28 & 58.07 & 46.22 & 62.83 & 42.47 & {\ul 43.49} & 61.67 & 38.23 & 54.21 \\
Gemini2.0 Flash & {\ul 57.41} & {\ul \textbf{76.46}} & {\ul 62.27} & {\ul \textbf{51.94}} & 75.12 & {\ul 47.59} & 34.85 & 68.62 & {\ul 44.54} & {\ul 59.82} \\
GPT-4o & 51.24 & 69.13 & 55.90 & 22.63 & 66.80 & 44.05 & 35.38 & 65.13 & 41.72 & 52.91 \\
GPT-4o mini & 43.92 & 60.41 & 53.81 & 28.17 & 56.37 & 36.36 & 30.40 & 52.86 & 33.25 & 46.12 \\ \midrule
Avg. & 48.72 & 61.54 & 53.18 & 36.64 & 58.17 & 41.75 & 36.08 & 60.60 & 37.84 & 49.68 \\
Human & -- & 75.33 & 77.75 & 23.81 & 88.75 & 58.33 & 48.50 & 90.00 & 55.25 & 64.72 \\ \bottomrule
\end{tabular}}
\end{table*}

\noindent\textbf{Human performance annotation.} After all task annotations are done, we further collect their human performance. We pick 30 samples from each dataset and assign these samples to annotators to generate answers. We use this answer to compute the human performance. We also ask for feedback and comments from the annotator. If the annotator raises any concerns about the dataset, we revise its construction until the samples resolve the annotator's concerns.

\subsection{Data Synthesis}
While real-world datasets provide a comprehensive evaluation of diverse scenes, a good diagnostic evaluation can point out the issues of VLMs, and further enhance the understanding of the model defects, guiding future research directions. To this end, we propose two diagnostic datasets with HRIs aiming to find the defect of VLMs on HRI understanding in two aspects.

\noindent\textbf{WhiteBackground NIAH.} This test aims to detect the regional defect of the VLMs, namely, Regional Divergence. Inspired by the NIAH test of long text~\cite{liu-etal-2024-lost} and the eye exam of humans, we propose to use a needle image to combine with a white background haystack to form an NxN grid (Figure~\ref{fig:example}). Specifically, we first prepare the image-question pairs from the VQAv2 dataset~\cite{goyal2017making} and use the image as the \textit{needle}. We then place it on different rows and columns of white girds as \textit{haystack} and evaluate the differences in model performance.

\noindent\textbf{ComplexGrid NIAH.} This test is to diagnose the VLM capability of retrieving the correct image among multiple distracting images. Also inspired by the NIAH test in long text~\cite{liu-etal-2024-lost}, we use an image search tool to extract the most similar images from the dev set of VQAv2. Then, we composite them to form a larger grid. We collect the caption of the needle and ask the model to point out the rows and columns where the needle is located (Figure~\ref{fig:example}).

\subsection{Data Analysis}
Table~\ref{tab:statis} and
Figure~\ref{fig:res_distribution} shows the statistics of the datasets. As shown in Figure~\ref{fig:res_distribution}, all of the datasets have a resolution higher than 1k. For most of the datasets, the resolution is between 1k and 8k while PANDA contains images with around $5\times10^8$ pixels, showing the diverse high-resolution distribution of \ours.   
Figure~\ref{fig:example} displays some examples of \ours, the questions of \ours~cover a wide range, such as indexing the correct image, action prediction, counting, and text recognition. More examples can be found in Supplementary materials.

\subsection{Data Preparation and Release}
\ours~consists of 7068 samples, divided into three splits: \textit{val}, \textit{testmini}, and \textit{test}. 
\textit{val} contains 750 samples. These samples are identical to human-annotated ones, designed for fine-grained validation of the users' VLM settings. 
\textit{testmini} comprises 1000 samples, picked from each \ours~real-world datasets, intended for rapid model development evaluation or for those with limited computing resources. To ensure \textit{testmini} maintains a distribution closely resembling the whole set, we adopt this sampling:  (1) first, randomly sample questions with a threshold number of 25 for each dataset; (2) then, randomly sample the remaining questions for each dataset according to its proportion in the entire set.
The \textit{test} features the remaining 5323 samples for standard evaluation. Notably,
the answer labels for \textit{test} will not be publicly released to facilitate fair evaluation. Instead, we
maintain an online evaluation platform for user submissions. Evaluation results will be shown on an official leaderboard.

\begin{table*}[]
\centering
\caption{Diagnostic NIAH test on WhiteBackground dataset, \textbf{bold} indicates the best performance.}
\label{tab:whitebg}
\resizebox{\linewidth}{!}{
\begin{tabular}{@{}crrrrrrrrrrrrr@{}}
\toprule
 & \multicolumn{1}{c}{1x1} & \multicolumn{3}{c}{3x3} & \multicolumn{3}{c}{5x5} & \multicolumn{3}{c}{7x7} & \multicolumn{3}{c}{10x10} \\ \cmidrule(lr){2-2}\cmidrule(lr){3-5}\cmidrule(lr){6-8}\cmidrule(lr){9-11}\cmidrule(lr){12-14} 
\multicolumn{1}{l}{} & \multicolumn{1}{l}{Perf $\uparrow$} & \multicolumn{1}{l}{Perf $\uparrow$} & \multicolumn{1}{l}{Size $\uparrow$} & \multicolumn{1}{l}{Region $\downarrow$} & \multicolumn{1}{l}{Perf $\uparrow$} & \multicolumn{1}{l}{Size $\uparrow$} & \multicolumn{1}{l}{Region $\downarrow$} & \multicolumn{1}{l}{Perf $\uparrow$} & \multicolumn{1}{l}{Size $\uparrow$} & \multicolumn{1}{l}{Region $\downarrow$} & \multicolumn{1}{l}{Perf $\uparrow$} & \multicolumn{1}{l}{Size $\uparrow$} & \multicolumn{1}{l}{Region 
 $\downarrow$} \\ \midrule

InterVL2 - 1B & 62.46 & 60.17 & \cellcolor[HTML]{FFFCF6}-2.29 & \cellcolor[HTML]{F1B2AD}25.16 & 55.72 & \cellcolor[HTML]{FFF8E8}-6.74 & \cellcolor[HTML]{EB928A}34.12 & 52.29 & \cellcolor[HTML]{FFF6DD}-10.17 & \cellcolor[HTML]{E77F76}39.37 & 53.11 & \cellcolor[HTML]{FFF6E0}-9.35 & \cellcolor[HTML]{F2B7B2}23.82 \\
InterVL2 - 2B & 65.66 & 63.91 & \cellcolor[HTML]{FFFDF8}-1.75 & \cellcolor[HTML]{F1B5B0}24.44 & 60.29 & \cellcolor[HTML]{FFFAEC}-5.37 & \cellcolor[HTML]{EC9B94}31.63 & 56.77 & \cellcolor[HTML]{FFF7E1}-8.89 & \cellcolor[HTML]{EA8E86}35.22 & 49.76 & \cellcolor[HTML]{FFF1CB}-15.90 & \cellcolor[HTML]{E67C73}40.03 \\
InterVL2 - 4B & 62.06 & 62.45 & \cellcolor[HTML]{FFFEFE}\textbf{0.39} & \cellcolor[HTML]{EFACA6}27.01 & 58.19 & \cellcolor[HTML]{FFFBF1}-3.87 & \cellcolor[HTML]{EA8D86}35.34 & 54.60 & \cellcolor[HTML]{FFF8E6}-7.46 & \cellcolor[HTML]{E78077}39.15 & 56.54 & \cellcolor[HTML]{FFF9EC}-5.52 & \cellcolor[HTML]{F5C7C3}19.45 \\
InterVL2 - 8B & 68.66 & 66.10 & \cellcolor[HTML]{FFFCF5}-2.56 & \cellcolor[HTML]{F1B4AF}24.69 & 62.66 & \cellcolor[HTML]{FFF9EA}-6.00 & \cellcolor[HTML]{ED9E97}30.87 & 58.86 & \cellcolor[HTML]{FFF6DE}-9.80 & \cellcolor[HTML]{E98B83}35.93 & 53.44 & \cellcolor[HTML]{FFF1CD}-15.22 & \cellcolor[HTML]{F3BCB8}22.42 \\
Phi - 4B & 77.46 & 74.00 & \cellcolor[HTML]{FFFBF2}-3.46 & \cellcolor[HTML]{FBE9E7}10.06 & 69.64 & \cellcolor[HTML]{FFF8E4}-7.82 & \cellcolor[HTML]{F6D0CD}17.02 & 65.71 & \cellcolor[HTML]{FFF4D8}-11.75 & \cellcolor[HTML]{F4C3BF}20.54 & 61.44 & \cellcolor[HTML]{FFF1CB}-16.02 & \cellcolor[HTML]{F3C0BC}21.28 \\
Qwen2-VL - 7B & 85.93 & 84.22 & \cellcolor[HTML]{FFFDF8}-1.71 & \cellcolor[HTML]{FEFAFA}5.30 & 83.14 & \cellcolor[HTML]{FFFCF4}-2.79 & \cellcolor[HTML]{FEF6F5}\textbf{6.52} & 81.71 & \cellcolor[HTML]{FFFBF0}-4.22 & \cellcolor[HTML]{FDF2F1}7.55 & 79.91 & \cellcolor[HTML]{FFF9EA}-6.02 & \cellcolor[HTML]{FBE7E6}10.56 \\
LLaMa3.2 - 11B & 72.53 & 69.14 & \cellcolor[HTML]{FFFBF2}-3.39 & \cellcolor[HTML]{F7D4D2}15.72 & 66.11 & \cellcolor[HTML]{FFF9E9}-6.42 & \cellcolor[HTML]{F5C6C2}19.75 & 61.73 & \cellcolor[HTML]{FFF5DB}-10.80 & \cellcolor[HTML]{EFABA5}27.13 & 41.17 & \cellcolor[HTML]{FFE49A}-31.36 & \cellcolor[HTML]{FFFFFF}\textbf{3.84} \\
InterVL2 - 26B & 82.60 & 80.87 & \cellcolor[HTML]{FFFDF8}-1.73 & \cellcolor[HTML]{FEF6F5}6.45 & 77.43 & \cellcolor[HTML]{FFFAED}-5.17 & \cellcolor[HTML]{FBE5E4}11.03 & 74.63 & \cellcolor[HTML]{FFF7E4}-7.97 & \cellcolor[HTML]{F9DDDB}13.29 & 65.41 & \cellcolor[HTML]{FFF0C7}-17.19 & \cellcolor[HTML]{F6D0CD}16.99 \\
InterVL2 - 40B & 84.53 & 83.42 & \cellcolor[HTML]{FFFDFA}-1.11 & \cellcolor[HTML]{FFFDFD}\textbf{4.57} & 80.02 & \cellcolor[HTML]{FFFAEF}-4.51 & \cellcolor[HTML]{FCEDEC}8.84 & 78.16 & \cellcolor[HTML]{FFF9E9}-6.37 & \cellcolor[HTML]{FAE5E3}11.29 & 74.95 & \cellcolor[HTML]{FFF6DF}-9.58 & \cellcolor[HTML]{F9DEDB}13.18 \\
InternVL2-Llama3-76B & 85.33 & 83.74 & \cellcolor[HTML]{FFFDF8}-1.59 & \cellcolor[HTML]{FFFDFD}4.59 & 77.09 & \cellcolor[HTML]{FFF7E3}-8.24 & \cellcolor[HTML]{FCEBE9}9.63 & \multicolumn{1}{l}{--} & \multicolumn{1}{l}{--} & \multicolumn{1}{l}{--} & 75.64 & \cellcolor[HTML]{FFF6DF}-9.69 & \cellcolor[HTML]{F9DDDB}13.35 \\
DeepSeek-VL2-27B & 72.06 & 49.71 & \cellcolor[HTML]{FFEBB7}-22.35 & \cellcolor[HTML]{F7D4D1}15.75 & 34.29 & \cellcolor[HTML]{FFDE86}-37.77 & \cellcolor[HTML]{F2B9B4}23.37 & 28.41 & \cellcolor[HTML]{FFD974}-43.65 & \cellcolor[HTML]{F0B0AB}25.72 & 23.95 & \cellcolor[HTML]{FFD666}-48.11 & \cellcolor[HTML]{F2B9B4}23.30 \\
LLava-Onevision-72B & \textbf{87.73} & \textbf{84.51} & \cellcolor[HTML]{FFFBF3}-3.22 & \cellcolor[HTML]{FFFBFA}5.14 & \textbf{84.04} & \cellcolor[HTML]{FFFBF2}-3.69 & \cellcolor[HTML]{FBE8E6}10.40 & \multicolumn{1}{l}{--} & \multicolumn{1}{l}{--} & \multicolumn{1}{l}{--} & \multicolumn{1}{l}{--} & \multicolumn{1}{l}{--} & \multicolumn{1}{l}{--} \\
Qwen2-VL - 72B & 84.13 & \textbf{84.51} & \cellcolor[HTML]{FFFEFE}0.38 & \cellcolor[HTML]{FEF9F9}5.62 & \textbf{84.04} & \cellcolor[HTML]{FFFEFD}\textbf{-0.09} & \cellcolor[HTML]{FEF5F5}6.62 & \textbf{84.20} & \cellcolor[HTML]{FFFEFD}\textbf{0.07} & \cellcolor[HTML]{FEF5F5}\textbf{6.65} & \textbf{84.56} & \cellcolor[HTML]{FFFFFF}\textbf{0.43} & \cellcolor[HTML]{FCEBE9}9.61 \\
LLaMa3.2 - 90B & 75.46 & 72.69 & \cellcolor[HTML]{FFFCF4}-2.77 & \cellcolor[HTML]{FBE6E4}10.88 & \multicolumn{1}{l}{--} & \multicolumn{1}{l}{--} & \multicolumn{1}{l}{--} & \multicolumn{1}{l}{--} & \multicolumn{1}{l}{--} & \multicolumn{1}{l}{--} & \multicolumn{1}{l}{--} & \multicolumn{1}{l}{--} & \multicolumn{1}{l}{--} \\ \midrule
GPT-4o-mini & 68.66 & 60.69 & \cellcolor[HTML]{FFF7E4}-7.97 & \cellcolor[HTML]{F9DCD9}13.77 & 52.53 & \cellcolor[HTML]{FFF1CA}-16.13 & \cellcolor[HTML]{F5C6C3}19.59 & 44.35 & \cellcolor[HTML]{FFEAB1}-24.31 & \cellcolor[HTML]{F0B1AB}25.64 & 32.94 & \cellcolor[HTML]{FFE08D}-35.72 & \cellcolor[HTML]{EB948C}33.65 \\
Claude-3-Haiku & 62.60 & 59.92 & \cellcolor[HTML]{FFFCF5}-2.68 & \cellcolor[HTML]{F6D0CD}17.00 & 54.05 & \cellcolor[HTML]{FFF7E2}-8.55 & \cellcolor[HTML]{F0ADA7}26.61 & 49.89 & \cellcolor[HTML]{FFF3D5}-12.71 & \cellcolor[HTML]{EC9A93}31.84 & 44.83 & \cellcolor[HTML]{FFEFC5}-17.77 & \cellcolor[HTML]{E98880}36.89 \\
Gemini-2.0-Flash & 69.86 & 67.45 & \cellcolor[HTML]{FFFCF6}-2.41 & \cellcolor[HTML]{FDF5F4}6.75 & 67.35 & \cellcolor[HTML]{FFFCF5}-2.51 & \cellcolor[HTML]{F7D3D0}16.00 & 66.50 & \cellcolor[HTML]{FFFBF3}-3.36 & \cellcolor[HTML]{F0B1AB}25.64 & 63.86 & \cellcolor[HTML]{FFF9EA}-6.00 & \cellcolor[HTML]{E77D74}39.85 \\ \bottomrule

\end{tabular}}

\end{table*}

\section{Experiments}
\subsection{Implementation Details}
We include a total of 28 VLMs in our experiment. The details are in supplementary materials. This includes one \textbf{Phi-3.5}~\cite{abdin2024phi}, two
\textbf{DeepSeek}~\cite{chen2025janus}, seven \textbf{InterVL2}~\cite{chen2024expanding,gao2024mini,chen2024far,chen2024internvl}, two \textbf{Qwen2-VL}~\cite{Qwen2VL}, two \textbf{MolMo}~\cite{deitke2024molmo}, one \textbf{LLaVA-Onevision}~\cite{liu2023visual},one \textbf{LLaVA-HR}~\cite{mra},  four \textbf{Llava-Next}~\cite{zhang2024video}, two \textbf{Llama-3.2}~\cite{llama}, and two \textbf{GPT-4o}~\cite{gpt4o}, two \textbf{Gemini}~\cite{team2024gemini}, and two \textbf{Claude}~\cite{anthropic2024claude}.

For WhiteBackground, we follow the accuracy in VQAv2\cite{goyal2017making}. For ComplexGrid dataset, we prompt the model to generate the column and row of the needle image and compare with the gold column and row using an exact match. For real-world datasets, since they are MCQ-based, we directly use exact math as metrics. Supplementary materials show more details.

\subsection{Overall Results on Real-world Datasets}
Table~\ref{tab:overall} shows the comparison of all models on \ours~real-world datasets. To make the comparison fair, we cluster the open-sourced models into 3 groups with similar parameter sizes (Small: 1B-13B, Medium: 14B-34B, Large: $>$35B), and compare the models inside each group. As can be seen, the best small model is Qwen2-VL 7B, with an average performance of 56.65\% on all 25 tasks. However, InternVL2 4B obtains group SOTA on Urban and Medical categories. For Medium, InternVL2 40B obtains the highest average score of 58.45\%. While for Large models, Qwen2-VL 72B obtains the best performance again, with 61.85\% on average. For proprietary models, the best model is Gemini2.0 Flash with 59.82\% performance. Supplementary materials contain the full results of all models.

\begin{figure*}[t]
    \centering
    \begin{subfigure}{0.32\textwidth}
        \centering
        \includegraphics[width=\linewidth]{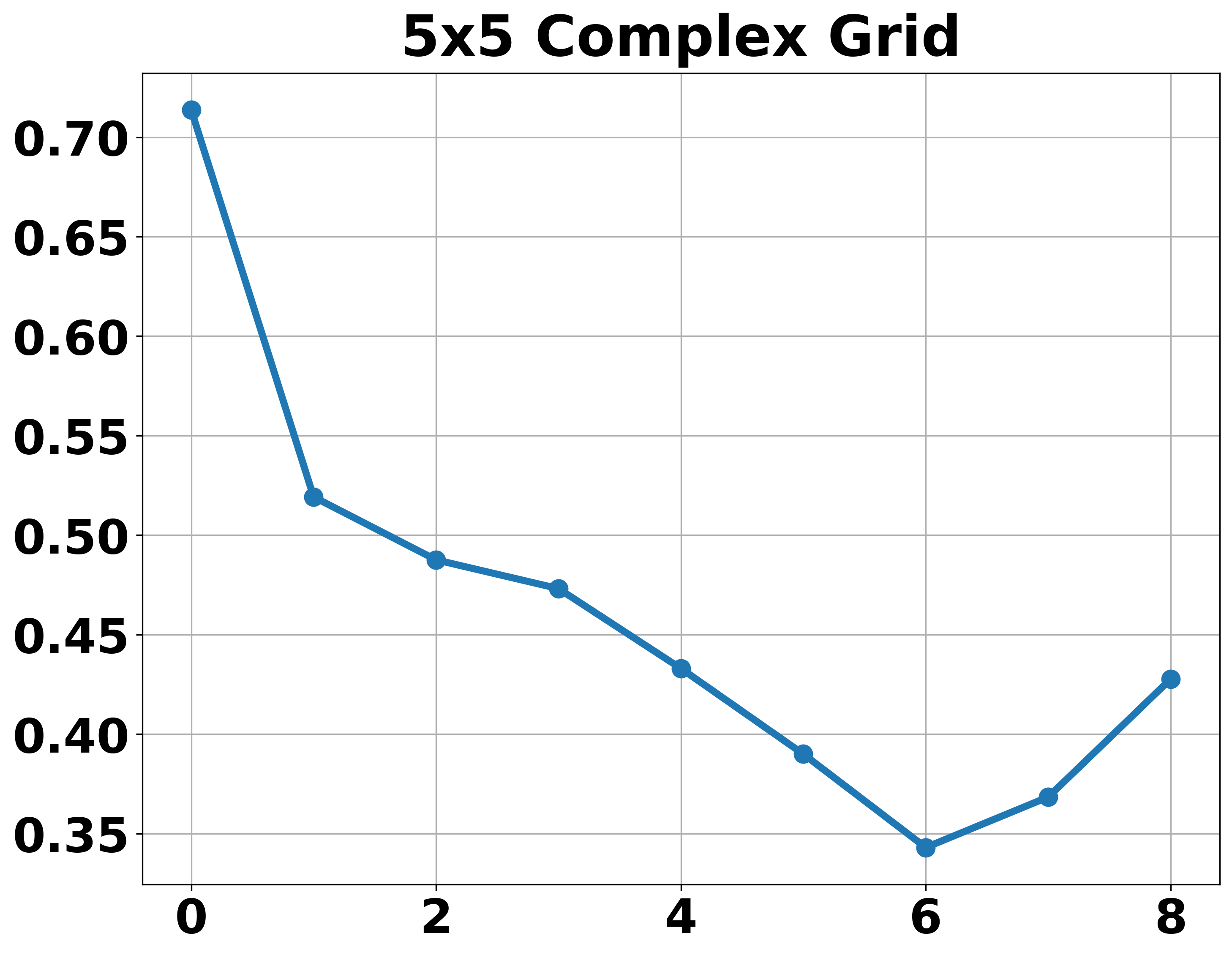}
       
    \end{subfigure}
    \hfill
    \begin{subfigure}{0.32\textwidth}
        \centering
        \includegraphics[width=\linewidth]{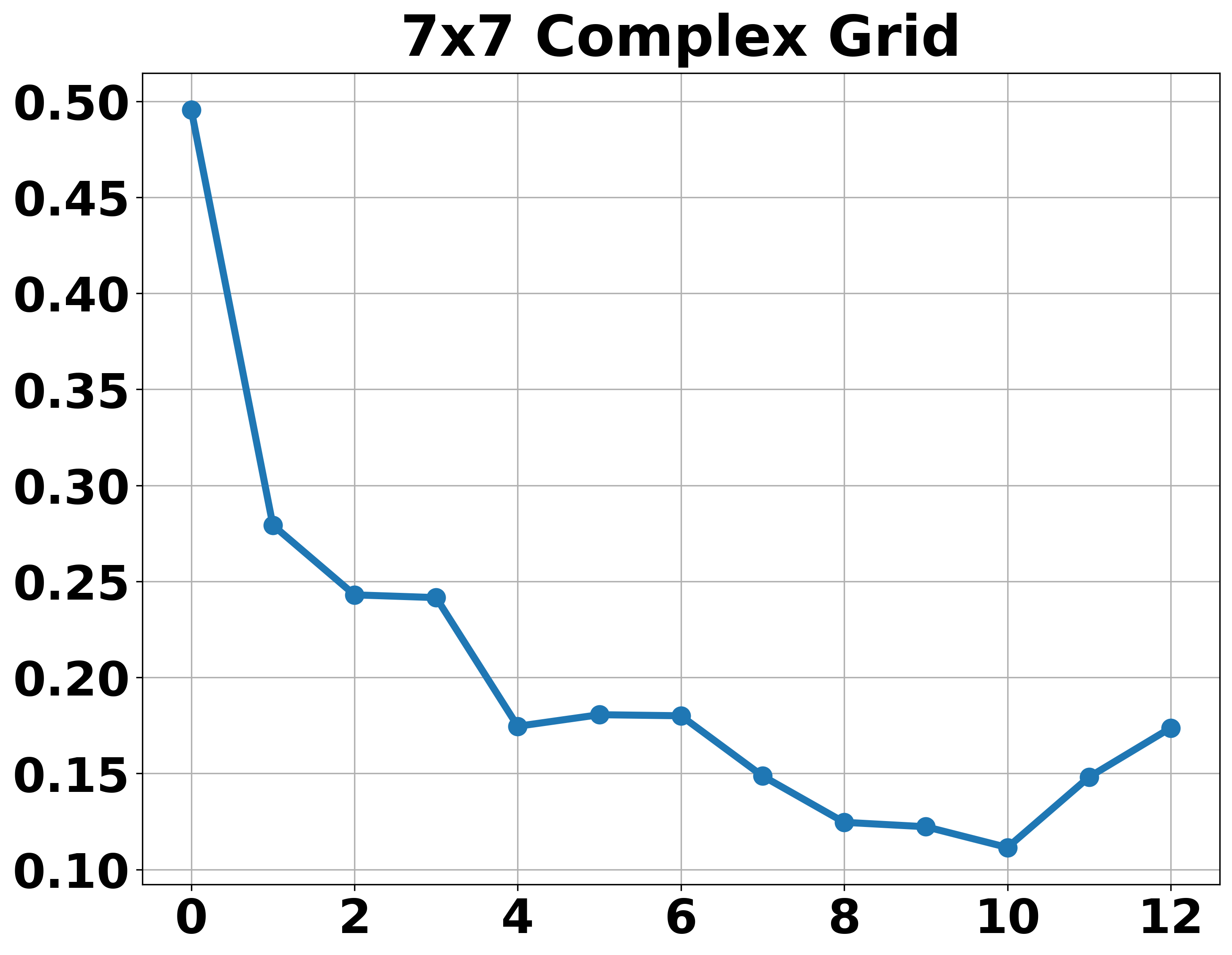}
      
    \end{subfigure}
    \hfill
    \begin{subfigure}{0.32\textwidth}
        \centering
        \includegraphics[width=\linewidth]{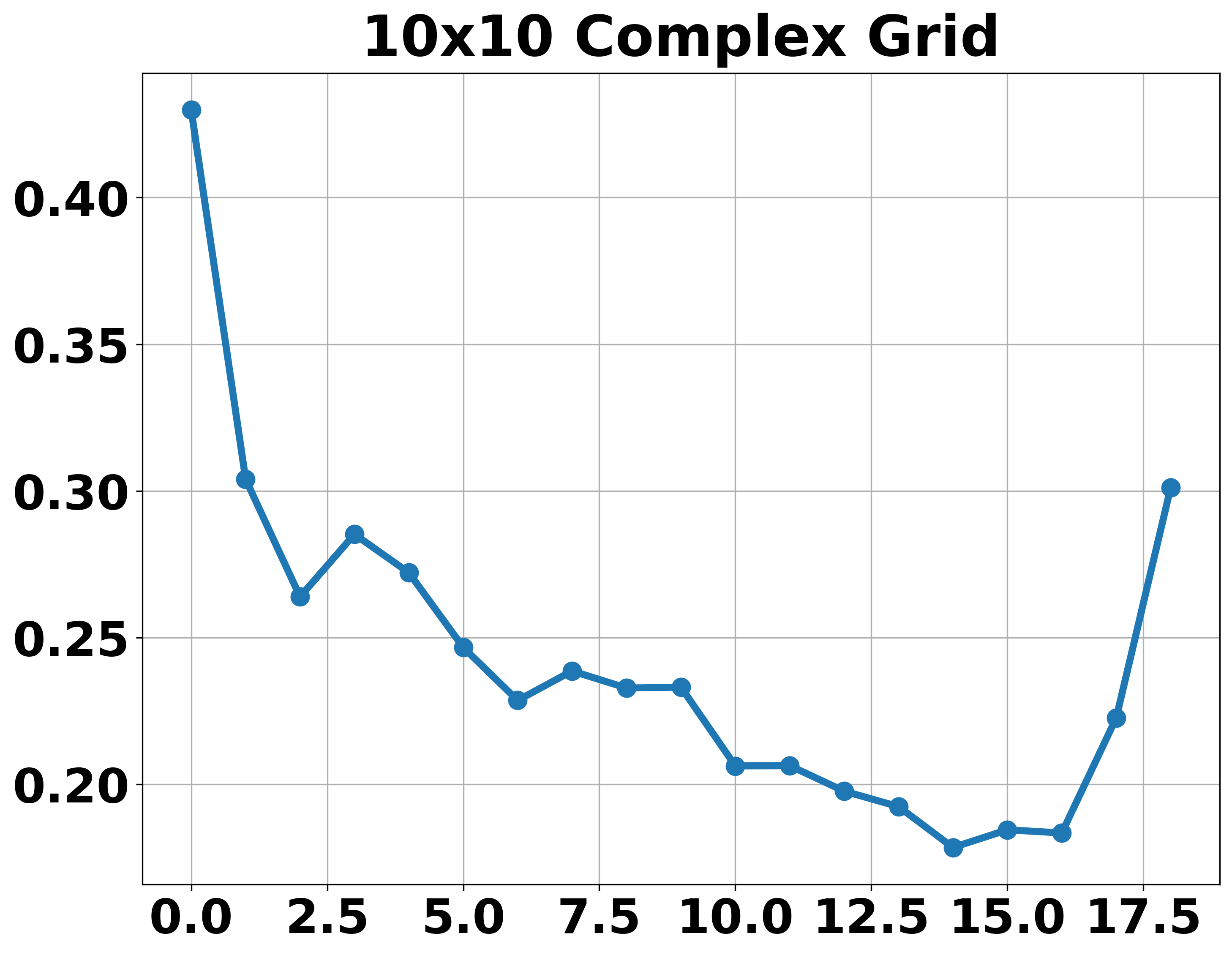}
       
    \end{subfigure}
    \caption{Performance of the regions averaged across all dataset points and all 18 VLMs. X-Axis is the Manhattan distance to the left upper corner, $|x-1| + |y-1|$ where $x,y$ is the row and column of the needle image, while the y-axis is the performance of that sample. With the increase of the x-axis, the performance of the model exhibits a U-shape, with much lower performance in the middle. With the increase in the image size, the shape becomes more significant. }
    \label{fig:lost}
\end{figure*}

Comparing these four types of models (Open-sourced Small, Medium, Large, and Proprietary), we can observe that the performance increases with the increasing model sizes. However, the best model globally is Qwen2-VL 72B, the only model whose performance is above 60\%, surpassing the GPT 4o, and Gemini. We further explore the reason by checking the dataset details. As shown in Table~\ref{tab:fine}, we find that GPT 4o outperforms Qwen2-VL 72B on HRVQA with 1k resolution, while Qwen2-VL performs much better on Galaxy with images as large as 20k resolution. Qwen can input images with native resolution, while GPT has a size limit of 5 MB. This shows that due to the native resolution support of Qwen, it obtains SOTA, even general capability might not be the best. This result highlights \textbf{the importance of the HRI processing capability of native resolution to obtain high performance. }

On the other hand, for VLMs, the difficulties of each category vary. As shown in the second-to-last row of Table~\ref{tab:overall}, Medical and Research obtain the lowest performance as it requires domain knowledge, such as pathology and medical images. Then, Remote and Urban planning is also challenging because it involves intensive counting and small object detection tasks. The simplest ones are Art and Sub-Img, which mainly focus on global understanding and reasoning capabilities. However, \textbf{the average performance across all categories is only 49.68\%, showing the large gap between VLMs and efficient HRI processing.}

\begin{table}[!ht]
\centering
\caption{Fine-grained comparison between Qwen 72B and GPT 4o on two datasets.}
\label{tab:fine}
\begin{tabular}{@{}lrr@{}}
\toprule
 & HRVQA (1k) & Galaxy (20k) \\ 
 \midrule
Qwen 72B & 69.59 & \textbf{80.80} \\ 
GPT 4o & \textbf{73.82} & 68.68 \\ \bottomrule
\end{tabular}
\end{table}
\begin{figure*}[t]
    \centering
    \begin{subfigure}{0.49\textwidth}
        \centering
        \includegraphics[width=\linewidth]{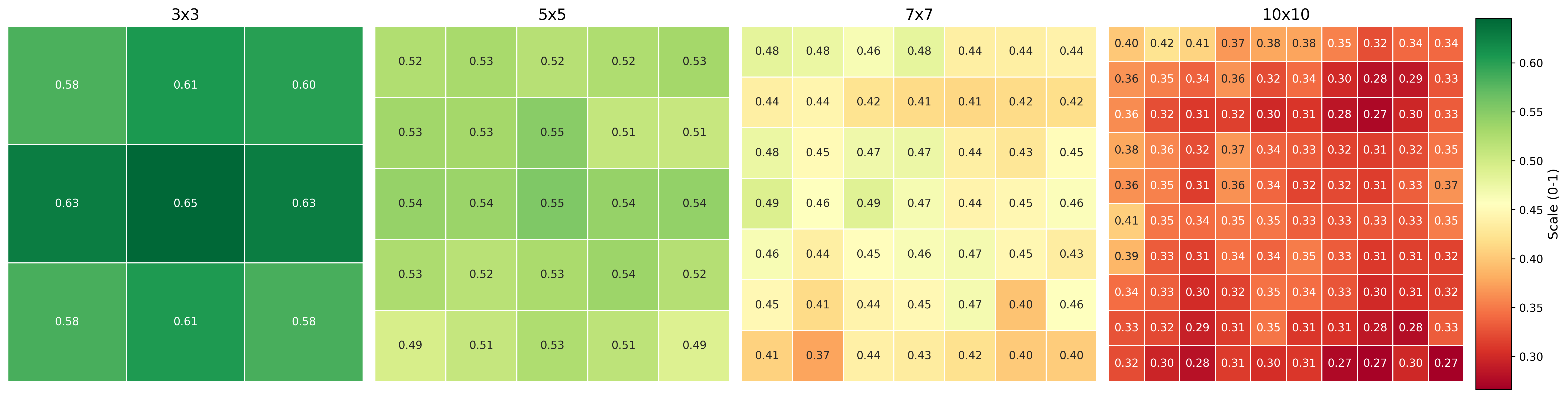}
        \caption{WhiteBackground, GPT-4o-mini.}
       
    \end{subfigure}
    \hfill
    \begin{subfigure}{0.49\textwidth}
        \centering
        \includegraphics[width=\linewidth]{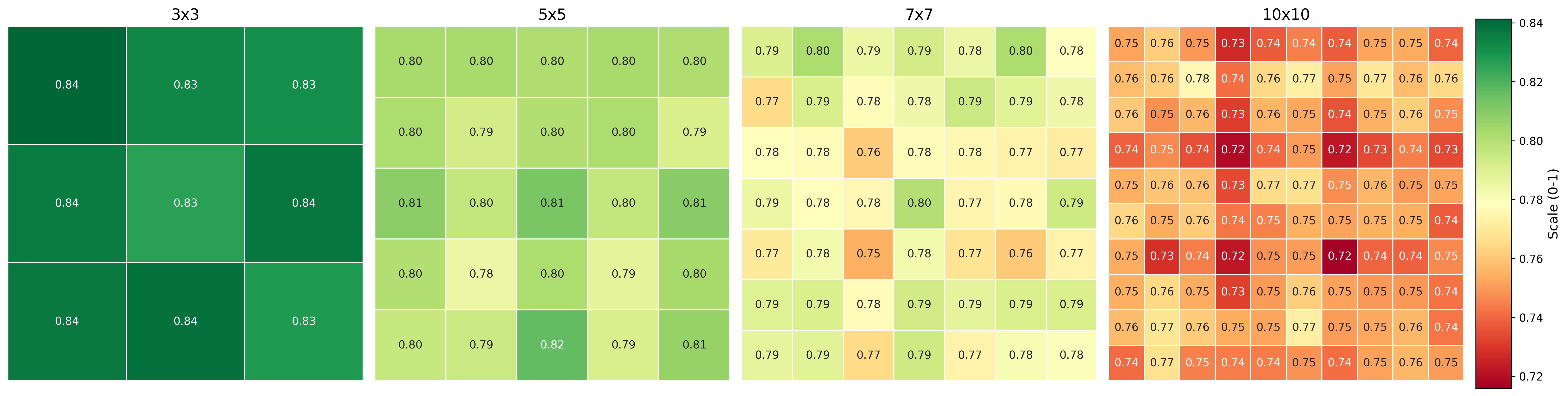}
        \caption{WhiteBackground, InternVL2-40B.}
       
    \end{subfigure}
    \hfill

    \begin{subfigure}{0.49\textwidth}
        \centering
        \includegraphics[width=\linewidth]{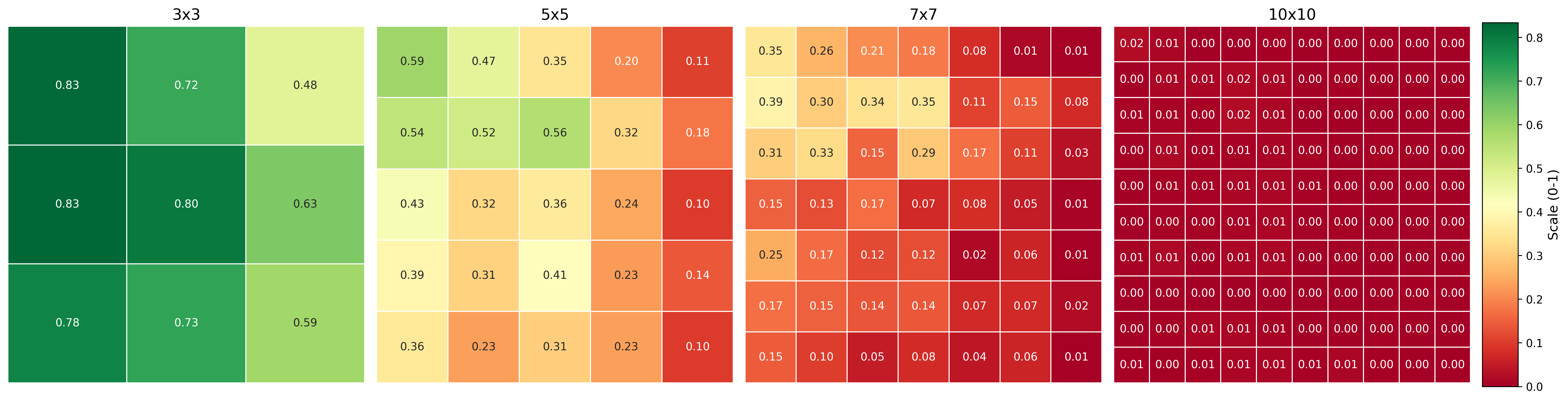}
        \caption{ComplexGrid, Claude-3.5-Haiku.}
        
    \end{subfigure}
    \hfill
    \begin{subfigure}{0.49\textwidth}
        \centering
        \includegraphics[width=\linewidth]{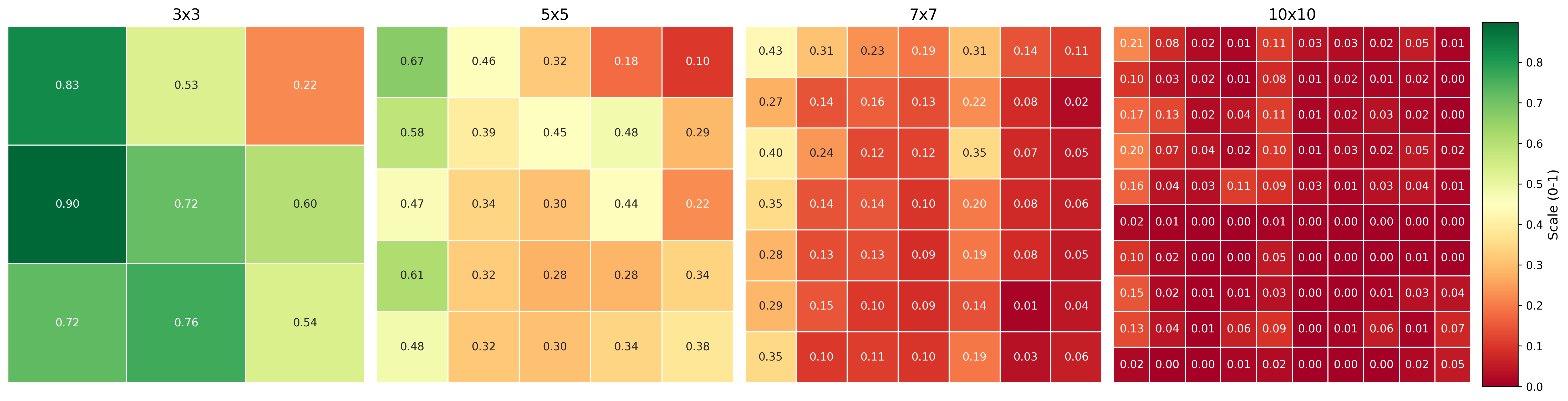}
        \caption{ComplexGrid, Phi-35-Vision.}
        
    \end{subfigure}
    \hfill

    \caption{Detailed performance of some models on two diagnose datasets.}
    \label{fig:heat}
\end{figure*}

\subsection{Overall Results on Diagnostic Datasets}
\paragraph{WhiteBackground.} Table~\ref{tab:whitebg} shows the statistics of the WhiteBackground diagnosis. We report the average performance of the samples (Perf $\uparrow$), the performance drop with image size increasing from 1x1 (Size $\uparrow$), and the region expectation gap (Region $\downarrow$), which is the difference between the highest performance region and the mean performance of every region. We call this \textit{Regional Divergence}. We propose to use these metrics because the model can be improved by (1) being more robust on image size extension, especially with simple white background fillings, and (2) inside each HRI, maintaining the same performance with each region, specifically being the same performance as the highest region. As shown in Table~\ref{tab:whitebg}, most of the models cannot maintain consistent performance with increasing image size. Furthermore, models exhibit significant Region Divergence, usually amplified with increasing image size. For instance, Gemini-2.0-Flash obtains 39.85\% Divergence on 10x10 grids, meaning that if the answer to the question is located at a random region, the performance will be around 40\% lower than the best one among 100 regions.

\begin{figure}[!ht]
    \centering
    \includegraphics[width=\linewidth]{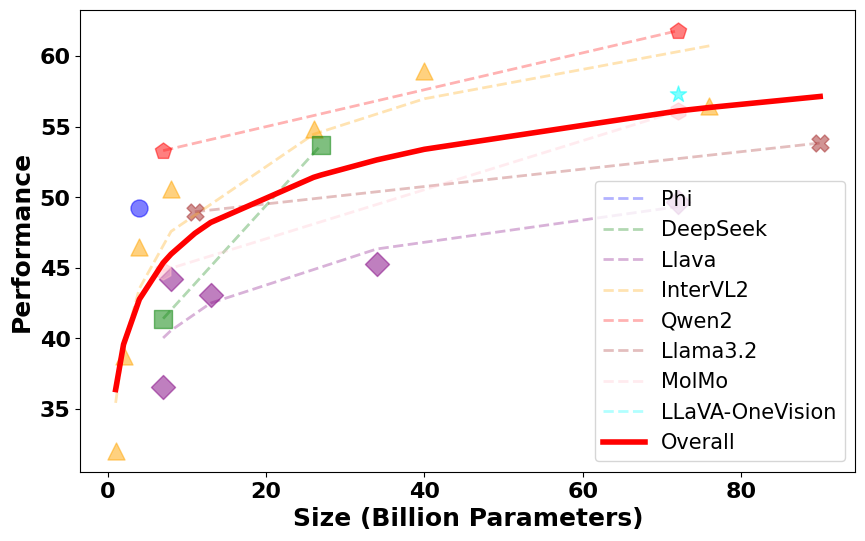}
    \caption{The relationship between model performance and model parameter size.}
    \label{fig:perf_size}
\end{figure}

\paragraph{ComplexGrid.} In the NIAH test~\cite{liu-etal-2024-lost}, the authors found a significant drop when the gold answer is in the middle of a long context, namely lost-in-the-middle. This is also observed in multi-modal settings when the image is mixed with text~\cite{song2024milebench}. Surprisingly, we discover a similar but non-identical behavior in HRI. 
Figure~\ref{fig:lost} shows the performance change of the models with increasing Manhattan distance from row 1, column 1 to the needle image. For instance, if the needle image is row 2, column 3, the Manhattan distance is computed as (2-1)+(3-1)=3. We observe a phenomenon that is similar to lost-in-the-middle. Differently, we observe the performance forms a U-shape based on the Manhattan distance from the left upper corner rather than the linear depth of the needle in traditional NIAH. We demonstrate that lost-in-the-middle-manhattan is novel and from the original lost-in-the-middle in supplementary materials.

\section{Analysis}

\subsection{Influence of Model Size}
We analyze the influence of model parameters on the performance. We plot the relation between VLMs' parameter size and average performance on 25 datasets. Next, we draw the trend line to fit the performance change. As shown in Figure~\ref{fig:perf_size}, although different families of models' performances are different, their trends are all log-like increasing. This shows that (1) increasing the model size can effectively increase the HRI understanding, especially for the small models; and (2) the effect of increasing size is slowing down.

\subsection{Global and Local Perception Trade-off}
\begin{figure}
    \centering
    \includegraphics[width=\linewidth]{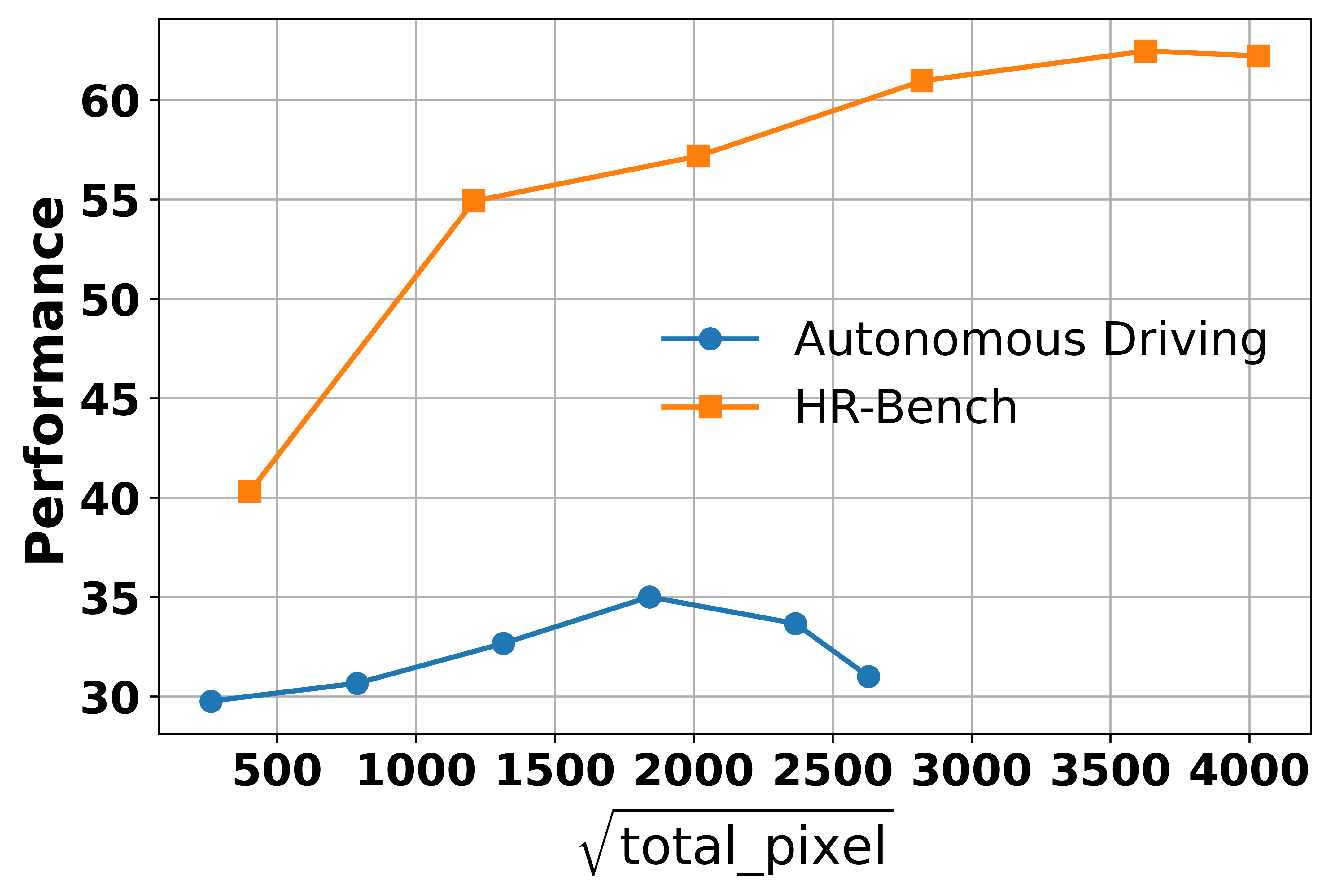}
    \caption{Performance change with image resizing on Qwen2-VL 7B.}
    \label{fig:trade-off}
\end{figure}

Processing HRIs requires capturing both fine-grained details and global context~\cite{hrsuvery}. When an image becomes larger, it becomes more difficult to capture global information because of the larger amount of patches in the input. At the same time, it is too difficult to recognize objects when the image becomes too small. To evaluate this trade-off in \ours, we run the Qwen2-VL 7B on two datasets: Autonomous Driving, and HR-Bench. We resize the image to 90\%, 70\%, 50\%, 30\%, and 10\% to evaluate the same image with different sizes. Results in Figure~\ref{fig:trade-off} show that resizing to 1840 can obtain the best performance of Autonomous Driving, and 3628 for HR-Bench. This \textbf{indicates the trade-off between local and global information can lead to an optimal point that is different in different }datasets. 

\subsection{Ablation on Multi-image Combination} 
For ComplexGrid dataset, multiple images are combined to form a large image with NxN grids. Thus, we conduct an ablation study on different combination methods to avoid unnecessary errors due to the non-optimal methods. We test 4 settings, dense is to combine all images without any interval between images or index text below each image. Then, we add an index to each image and a white interval between images. We evaluate four models on the 10x10 dataset~\ref{fig:example}. Results are in Table~\ref{tab:ablation}. Results show that indexing and interval are useful for the model to recognize the different sub-images and avoid errors in counting the index. Thus, we use both to construct the ComplexGrid dataset. 
\begin{table}[!ht]
\caption{Ablation study on different image combination methods.}

\label{tab:ablation}
\centering
\begin{tabular}{@{}lrrrr@{}}
\toprule
 & \multicolumn{2}{c}{Qwen2-VL} & \multicolumn{2}{c}{InterVL2} \\ \cmidrule(lr){2-3}\cmidrule(lr){4-5}
 & 7B & 72B & 8B & 26B \\ \midrule
dense & 1.53 & 3.17 & 2.78 & 1.82 \\
w/ index & 8.22 & 18.33 & 17.15 & 6.30 \\
w/ interval & 2.33 & 5.17 & 4.28 & 1.99 \\
w/ both & \textbf{8.62} & \textbf{18.72} & \textbf{20.54} & \textbf{8.28} \\ \bottomrule
\end{tabular}
\end{table}

\subsection{Fine-grained Analysis of Diagnostic Datasets}
To further analyze the performance details of two diagnostic datasets, we draw a heatmap for the models. Figure~\ref{fig:heat} shows the performance divergence on different regions, where each grid represents the performance when the needle image is on that grid. The results show that for the WhiteBackground dataset, the performance of different grids varies. Although different models do not have a unified pattern, the Regional Divergence is still significant, especially on larger images. For ComplexGrid, the results clearly show the lost-in-middle phenomenon with the increasing Manhattan distance, where the performance is the best at the upper left corners and gradually becomes worse with increasing Manhattan distance.

\section{Conclusion and Future Work}
In this paper, we propose \ours, a unified benchmark for HRI understanding, consisting of 25 real-world datasets and two diagnostic datasets. Results show that the models exhibit low performance on real-world tasks, showing the challenge of \ours, and display regional divergence and lost-in-the-middle on diagnostic datasets that can direct future improvement.
In the future, researchers can develop high-performance, general-purpose HRI processors by fine-tuning on synthetic datasets or explore the deeper reasons for utilization issues. After developing a stronger model, it can be tested on real-world datasets of \ours~and submitted to the leaderboard of \ours~to obtain a direct comparison with other models on real-world scenarios. 
{
    \small
    \bibliographystyle{ieeenat_fullname}
    \bibliography{main}
}

\clearpage
\setcounter{page}{1}
\maketitlesupplementary

\section{Dataset Details}
\label{sec:dataset_details}
\noindent\textbf{Autonomous Driving} We extract samples from the \textit{MME-Realworld} dataset to evaluate a model's embodied intelligence, focusing on perception tasks such as distant object perception, attribute recognition, and counting, as well as reasoning tasks including intention prediction, interaction relation understanding, and driver attention prediction.

\noindent\textbf{Monitoring} Extracted from \textit{MME-Realworld}, this dataset features images taken from public safety cameras in diverse scenarios. It features realworld challenges including varying object scales and partially out-of-view objects captured from different view points across day and night.

\noindent\textbf{Document Parsing} For text recognition in images, we adopt \textit{DocStruct4M}, which focuses on structure-aware parsing of complex document data in images across five domains: documents, webpages, tables, charts, and natural images. 

\noindent\textbf{Fine-grained Perception} We select \textit{HR-Bench} for fine-grained perception in high-resolution images. It poses single-instance and cross-instance perception tasks. The dataset is available in two resolution versions (4K and 8K), with the 8K images cropped around the relevant objects to produce the 4K versions. We select samples from both versions.

\noindent\textbf{Aerial Images}  \textit{HRVQA} is selected for aerial image understanding, it features images of a 1K spatial resolution and QA pairs that span 10 question types (Number, Yes/No, etc.) and 27 category concepts (Vehicles, Urban area, Water bodies, etc.).

\noindent\textbf{Image Quality} To evaluate models on quality assessment of daily-life pictures, we select \textit{HRIQ} designed for Blind Image Quality Assessment (BIQA) based on human perceivable factors like blur, exposure, noise, etc. The label is a human aligned Mean Opinion Score (MOS) on a scale of 0 to 5 given as options. We also design a custom prompt to instruct the model about the task and the response format.

\noindent\textbf{Infographics} Infographics contain a mix of textual and visual elements arranged in complex layouts. We leverage sampels from \textit{InfographicVQA} to test a model's ability to recognize and jointly reason over multiple spans of information present in infographics.

\noindent\textbf{Tissue Diagnosis} Automatic analysis of tissue samples can accelerate clinical diagnosis and treatment. To do this,we extract samples from \textit{LungHist700}, a collection of histopathological lung tissue images for the classification of lung malignancies. We design a custom prompt to instruct the model on the task, the options (seven classes), and the response format.

\noindent\textbf{Multi-Image} We choose \textit{MuirBench} for its diverse tasks and multi-image relationships. To enable a single high-resolution image input, we combine multiple images in each sample into a grid on a canvas. In addition, we select only samples with answers and remove any unanswerable questions.

\noindent\textbf{Chart Comprehension} Applying Large Multimodal Models (LMMs) to charts enables efficient information processing and extraction of insights. Although we have collected chart data from other datasets, we select \textit{NovaChart} for its comprehensiveness, featuring 18 different chart types and 15 chart-related tasks.

\noindent\textbf{Visual Difference} Describing differences between image sets is crucial in many real-world applications (cite). We repurpose \textit{VisDiffBench} by selecting smaller subsets of 20 samples from the original image sets and creating a high-resolution image as a 4x10 grid, with the first two rows occupied by images from set 1 and the last two rows by images from set 2.

\noindent\textbf{Medical Image} A VQA dataset for radiology images of various types (X-rays and CT scans) covering the chest and abdominal regions with diverse question about size, modality, abnormality, etc.

\noindent\textbf{Telescope Image} The \textit{Galaxy} dataset we use contains the images captured by a bubble telescope. We annotate this dataset from scratch with a question and four options.

\noindent\textbf{CAD} Contains floor-plan drawings of various architecture projects including residential buildings, schools, hospitals, and offices. It shows high variance in style and appearance of objects or symbols.

\noindent\textbf{PANDA} This dataset features high-resolution images with a wide Field-of-View (FoV) in outdoor scenarios, capturing pedestrians with varying crowd densities, poses, trajectories, and occlusions.

\noindent\textbf{V*} A dataset for testing models on perceiving small details in High-Resolution images of real-life scenarios. Sub-tasks include attribute identification and spatial relationship reasoning of small very small objects. 

\noindent\textbf{MileBench} We extract samples from MileBench, which evaluates multi-modal long-context understanding involving multiple images. The model must retain and integrate contextual information from extended inputs to answer questions accurately. The subtasks feature images that are temporally or semantically related.

\noindent\textbf{OCR in the Wild} Text recognition in real-world outdoor scenarios, such as streets and shops, involving the perception of advertisements, signage, identity information, and other textual elements. The samples are extracted from \textit{MME-Realworld}.

\noindent\textbf{Remote Sensing} Extracted from \textit{MME-Realworld}, this tests perception in high-resolution images with rich details, encompassing object counting, color recognition, and spatial relationship understanding.

\noindent\textbf{Chart and Diagram} Unlike other chart datasets, this dataset presents highly complex chart data, such as financial reports, which feature extensive numerical information and mathematical content. It evaluates both perception and reasoning capabilities of models. The perception tasks involve locating values in diagrams and tables, while the reasoning tasks include identifying maximum and minimum values, performing calculations, and predicting trends. The samples are extracted from \textit{MME-Realworld}.

\noindent\textbf{ArtBench} Contains artwork from 10 different artistic styles: Baroque, Surrealism, Post Impressionism, Realism, Romanticism, Impressionism, Art Nouveau, Expressionism, Renaissance, and Ukiyo-e. The correct artistic style along with few other distractor choices are given options.

\noindent\textbf{Museum} To assess models on art understanding, the \textit{MAMe} dataset comprises of various artworks and their corresponding medium (the various materials and techniques used to create the artwork). The dataset exhibits high intra-class variance, requiring models to pay close attention to fine-grained details.

\noindent\textbf{Animals} This dataset presents the task of counting various types of waterfowl using high-resolution aerial images of water bodies. This task is relevant for surveying waterfowl and reduces the manual effort.

\noindent\textbf{Product Anomaly Detection} Evaluating LMMs on their ability to identify defective (anomalous) products presents a highly industry-relevant task. This dataset not only supports anomaly detection but also includes additional subtasks for anomaly analysis, such as defect type classification, defect localization, and severity assessment.

\noindent\textbf{Grass} Automated inspection of vegetation, such as signal grass (\textit{Urochloa}), is crucial for farmers and promotes sustainable agriculture. To assess models in this real-world application, we adopt the task of phenological stage classification and raceme counting in high-resolution RGB images of \textit{Urochloa}.

\noindent\textbf{Diagnosis Datasets} For WhiteBackground dataset, we first pick 500 samples from VQAv2 dataset. Then, we combine each sample with white background images of different sizes. In this paper, we include 1x1 (no white background), 3x3, 5x5, 7x7, and 10x10 versions. NxN indicates the needle image is combined with $N\times N - 1$ white background images of the same size to form the entire image. In this case, the needle image has $N\times N$ positions for each sample. We run experiments to observe the difference in performance in each position and measure Regional Divergence.  Similarly, in ComplexGrid, we use similar images to fill the background rather than white background images. To pick out the most similar images, we use BLIP~\cite{https://doi.org/10.48550/arxiv.2201.12086} to rank the similarity between the needle image and all images in the validation set of VQAv2. And use the most similar $N\times N -1$ images as the haystack.

\begin{table*}[]
\caption{Overview of 25 real-world datasets and their statistics. * indicates that the dataset is reannotated.}
\label{tab:real_detail}
\resizebox{\linewidth}{!}{
\begin{tabular}{@{}lllrlll@{}}\toprule
Dataset Name & Explaination & Capability & \multicolumn{1}{l}{\# Samples} & Min Res & High Res & Avg. Res \\ \midrule
Autonomous\_Driving & Street View & Small Object Understanding & 300 & 5760x1200 & 5760x1200 & 5760x1200 \\
DocStruct4M* & Text document & OCR & 296 & 1024x1024 & 4000x28990 & 1733x2675 \\
HR-Bench & Daily photos & Small Object Understanding & 397 & 4032x1152 & 7680x7680 & 5740x4458 \\
HRVQA* & Aerial Image & Spactial relation, Small Objects & 273 & 1024x1024 & 1024x1024 & 1024x1024 \\
HRIQ & Long range picture & Small Object Understanding & 300 & 2880x2160 & 2880x2160 & 2880x2160 \\
InfographicVQA* & Graophic layout document & OCR & 300 & 1024x1024 & 6250x9375 & 1970x3881 \\
LungHist700 & Microscope Medical Image & Domain Knowledge & 308 & 1600x1200 & 1600x1200 & 1600x1200 \\
MuirBench* & Multi-imge combination & Multi-image Reasoning & 300 & 1064x1204 & 16062x7704 & 3072x2334 \\
NovaChart* & Chart Image & Chart Understanding & 297 & 2000x1600 & 3000x2147 & 2006x1600 \\
Video\_Monitoring & Street monitor & Small Object Understanding & 300 & 1280x1024 & 2048x2048 & 1989x1460 \\
VisDiffBench* & Multi-image combination & Multi-image Reasoning & 150 & 5220x1648 & 5220x2088 & 5220x2085 \\
VQA-RAD & Medical x-ray image & Domain Knowledge & 225 & 1024x1024 & 2321x1384 & 1041x1230 \\
Galaxy* & Telescope Image & Counting & 87 & 1435x732 & 29566x14321 & 4828x4078 \\
OCR\_in\_the\_Wild & Street brands & Small Object OCR & 300 & 1056x1056 & 7680x4320 & 2282x1867 \\
Remote\_Sensing & Shop signs & Small Object Understanding & 300 & 1272x1419 & 11500x7500 & 5788x4536 \\
Diagram\_and\_Table & Chart inside large image & Small Chart Object Understanding & 300 & 1201x1086 & 2481x3507 & 2337x1521 \\
VStar\_Bench & Daily photos & Image Search & 232 & 1080x1439 & 7500x5000 & 2357x1683 \\
MAME & Museum artiwork & Domain Knowledge & 300 & 1109x1043 & 15649x8900 & 3124x3200 \\
Izembek & Remote sensing of Zoo & Counting & 300 & 8688x5792 & 8688x5792 & 8688x5792 \\
ArtBench & Scanned Painting & Domain Knowledge & 306 & 1083x1024 & 9449x6496 & 1982x2017 \\
Grass & Argiculture Image & Counting & 300 & 4224x3168 & 4224x3168 & 4224x3168 \\
MMAD & Daily photo & Reasoning & 300 & 1024x1024 & 3024x3024 & 1918x1777 \\
MileBench & Video frame & Image Reseasoning & 300 & 1600x800 & 6400x6400 & 3096x2506 \\
PANDA & Public Monitor for Crowd & Crowd Counting & 300 & 24853x13983 & 35503x26627 & 27002x16152 \\
CAD* & Interior Design & Spactial relation, Counting & 297 & 2000x2000 & 2000x2000 & 2000x2000 \\
Total & N/A & N/A & 7068 & 1024x1024 & 35503x26627 & 5359x5395 \\ \bottomrule
\end{tabular}}
\end{table*}

\section{Ablation on Lost-in-the-Middle}
\label{sec:ablation_lost}
To test whether the observed U-shape is a trivial extension of existing work~\cite{liu-etal-2024-lost}, we further evaluate the model with flattened distance, the metric used in the original lost-in-the-middle that measures the linear distance of the starting token and needle tokens in the input. Since VLMs use a vision transformer~\cite{Qwen2VL} that inputs the image as linear patches, similarly, we measure the distance by counting how many patches the needle is from the first patch.

\begin{figure}
    \centering
    \includegraphics[width=\linewidth]{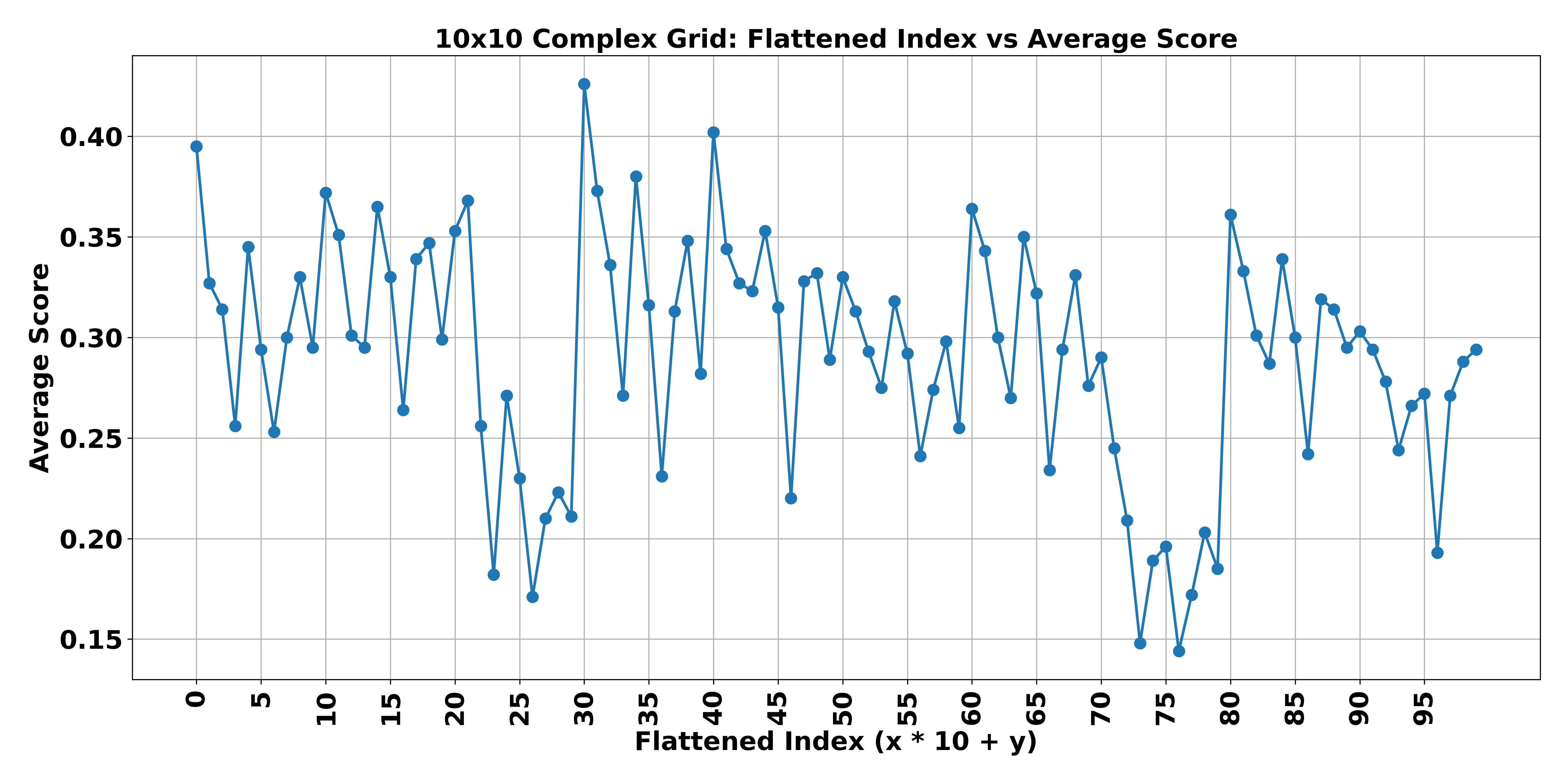}
    \caption{The performance of all models with an increase of patch id. Unlike lost-in-the-middle, no significant pattern can be observed.}
    \label{fig:flatten_lost}
\end{figure}

The results are shown in Figure~\ref{fig:flatten_lost}. As shown, no significant patter can be observed with the increasing of the patch distance, showing that the proposed phenomenon is not the same as the original lost-in-the-middle.

\section{Model Details}
\label{sec:model_details}
We include a total of 28 models in our experiment. \textbf{Phi-3.5}~\cite{abdin2024phi} is a lightweight model designed for efficient language understanding and generation. We include \textbf{Phi 3.5 vision instruct}~\cite{abdin2024phi} for experiments. \textbf{DeepSeek Janus Pro 7B}~\cite{chen2025janus} is a model that integrates multi-modal reasoning capabilities. \textbf{DeepSeek-VL2}~\cite{wu2024deepseek} is a vision-language model, with \textbf{deepseek vl2 27B} included in our evaluation. \textbf{InterVL2}~\cite{chen2024expanding,gao2024mini,chen2024far,chen2024internvl} is a family of multi-modal models ranging from small to large-scale by OpenGVLab. We include \textbf{InterVL2 1B}, \textbf{InterVL2 2B}, \textbf{InterVL2 4B}, \textbf{InterVL2 8B}, \textbf{InterVL2 26B}, \textbf{InterVL2 40B}, and \textbf{InterVL2 Llama3 76B} for experiments. \textbf{Qwen2-VL}~\cite{Qwen2VL} is a vision-language model, and we consider both \textbf{Qwen2 VL 7B Instruct} and \textbf{Qwen2 VL 72B Instruct}. \textbf{MolMo}~\cite{deitke2024molmo} is a series of models designed for molecular and scientific applications. We include \textbf{Molmo 72B 0924} and its distilled variant, \textbf{Molmo 7B D 0924}. \textbf{LLaVA-Onevision}~\cite{liu2023visual} is an open-source multimodal LLM, we selected \textbf{llava-onevision-qwen2-72b-ov-hf} model for our experiments. \textbf{Llava-Next}~\cite{zhang2024video} is an evolution of LLaVA, and we include \textbf{llama3-llava-next-8b-hf}, 
\textbf{llava-v1.6-vicuna-13b-hf}, \textbf{llava-v1.6-34b-hf}, and\textbf{ llava-next-72b-hf} in our experiments. \textbf{Llama3.2} builds on the Llama architecture with enhanced scalability. We include 
\textbf{Llama-3.2-11B-Vision-Instruct} and \textbf{Llama-3.2-90B-Vision-Instruct} in our experiments. \textbf{GPT}~\cite{achiam2023gpt}includes versions optimized for both efficiency and performance, with \textbf{GPT 4o} and \textbf{GPT 4o-mini} selected. Gemini is a family of LLMs, and we evaluate \textbf{Gemini 2.0 Flash} and \textbf{Gemini 1.5 Pro}~\cite{team2024gemini}. \textbf{Claude} is a family of LLMs known for its strong reasoning and safety features. We include two models in ascending order of capability: \textbf{Claude-3-haiku} and \textbf{Claude-3.5-sonnet}.

\section{Performance Details on Real-world Datasets}
\label{sec:perf_details}
Table~\ref{tab:all_perf_1} and Table~\ref{tab:all_perf_2} display the performance of all VLMs on every real-world dataset. The scores are the average performance of all samples in val, test, testmini splits. 

\begin{landscape}
\begin{table}[]
\centering
\caption{Performance of all VLMs on every real-world dataset (Part 1). }
\label{tab:all_perf_1}
\resizebox{\textwidth}{!}{%
\begin{tabular}{@{}ccccccccccccc@{}}
\toprule
 & \textbf{Drive} & \textbf{DocStr} & \textbf{HR-B} & \textbf{HRVQA} & \textbf{HRIQ} & \textbf{InfoQ} & \textbf{Lung} & \textbf{Muir} & \textbf{Nova} & \textbf{Monitor} & \textbf{VisDiff} & \textbf{RAD} \\ \midrule
\multicolumn{1}{c}{Random} & 20.00 & 25.02 & 25.00 & 25.06 & 20.00 & 25.06 & 14.29 & 23.38 & 23.70 & 20.00 & 25.00 & 33.33 \\
Calude3 Haiku & 27.35 & 62.05 & 29.43 & 62.32 & 29.15 & 56.14 & 14.11 & 50.00 & 57.21 & 19.40 & 84.09 & 46.20 \\
Calude3.5 Sonnect & 25.21 & 79.46 & 48.42 & 69.57 & 36.77 & 83.33 & 23.24 & 65.68 & 81.22 & 28.02 & 92.05 & 63.92 \\
Gemini1.5 Pro & 29.49 & 63.39 & 59.18 & 73.91 & 39.01 & 62.28 & 32.37 & 57.20 & 73.80 & 29.74 & 75.00 & 65.19 \\
Gemini2.0 Flash & 38.03 & 67.41 & 64.56 & 74.88 & 43.05 & 88.16 & 46.89 & 63.56 & 68.12 & 34.91 & 82.95 & 58.86 \\
DeepSeek-VL2 27B & 37.18 & 58.48 & 61.08 & 54.59 & 34.53 & 65.79 & 15.35 & 61.02 & 66.81 & 43.10 & 87.50 & 63.92 \\
GPT-4o & 32.05 & 67.86 & 55.70 & 72.95 & 44.39 & 74.12 & 1.24 & 58.47 & 73.80 & 34.91 & 89.77 & 51.90 \\
GPT-4o mini & 30.77 & 54.91 & 47.78 & 74.40 & 46.19 & 61.40 & 26.56 & 42.80 & 64.63 & 25.86 & 65.91 & 30.38 \\
InternVL2 1B & 22.22 & 35.27 & 38.92 & 36.23 & 21.97 & 36.84 & 12.45 & 30.51 & 32.31 & 21.98 & 28.41 & 54.43 \\
InternVL2 2B & 38.89 & 42.41 & 42.41 & 64.73 & 34.08 & 51.32 & 12.45 & 30.51 & 54.15 & 26.29 & 23.86 & 60.13 \\
InternVL2 4B & 35.90 & 56.25 & 46.84 & 48.79 & 29.15 & 63.60 & 19.09 & 58.90 & 58.95 & 35.78 & 85.23 & 73.42 \\
InternVL2 8B & 37.61 & 67.86 & 49.05 & 60.87 & 39.46 & 71.49 & 18.67 & 52.54 & 58.08 & 28.02 & 89.77 & 70.25 \\
InternVL2 26B & 42.74 & 68.30 & 60.44 & 58.45 & 32.74 & 70.61 & 17.01 & 58.90 & 62.45 & 40.09 & 85.23 & 67.09 \\
InternVL2 40B & 37.61 & 72.77 & 66.14 & 67.63 & 36.32 & 84.21 & 13.69 & 64.83 & 68.56 & 39.22 & 95.45 & 71.52 \\
InternVL2 76B & 38.46 & 69.64 & 59.81 & 58.45 & 39.46 & 84.65 & 14.11 & 65.68 & 55.90 & 41.81 & 94.32 & 70.89 \\
DeepSeek-Janus 7B & 30.34 & 39.73 & 31.96 & 51.69 & 27.35 & 41.67 & 14.94 & 48.31 & 41.92 & 22.41 & 51.14 & 64.56 \\
Llama3.2 11B & 32.91 & 58.93 & 52.85 & 70.05 & 21.52 & 67.11 & 25.73 & 46.19 & 63.32 & 30.60 & 81.82 & 63.92 \\
Llama3.2 90B & 31.62 & 72.77 & 54.11 & 74.88 & 23.32 & 71.05 & 17.01 & 54.66 & 69.00 & 34.48 & 82.95 & 71.52 \\
Llava-HR 7B & 31.20 & 28.57 & 41.46 & 55.07 & 28.70 & 37.72 & 14.52 & 25.85 & 38.86 & 31.90 & 47.73 & 44.94 \\
Llava-Next 8B & 34.19 & 41.96 & 44.62 & 67.15 & 21.97 & 47.81 & 12.45 & 40.25 & 43.23 & 28.45 & 79.55 & 60.76 \\
Llava-Next 13B & 28.63 & 48.21 & 40.82 & 54.59 & 31.39 & 46.05 & 13.28 & 45.76 & 55.46 & 37.50 & 80.68 & 68.35 \\
Llava-Next 34B & 29.91 & 66.96 & 53.80 & 66.18 & 36.32 & 59.65 & 15.77 & 58.90 & 65.50 & 31.90 & 80.68 & 65.82 \\
Llava-Next 72B & 29.91 & 67.41 & 51.58 & 63.77 & 34.08 & 63.16 & 14.94 & 46.61 & 64.63 & 35.78 & 85.23 & 65.19 \\
Llava-OneVision 72B & 32.91 & 72.32 & 62.34 & 68.12 & 46.64 & 80.70 & 15.35 & 66.10 & 69.87 & 33.19 & 75.00 & 78.48 \\
Phi3.5 4B & 32.91 & 54.02 & 45.89 & 65.22 & 43.50 & 62.28 & 7.05 & 46.19 & 58.08 & 31.03 & 89.77 & 67.72 \\
MolMo 7B-D & 33.76 & 52.68 & 46.52 & 55.07 & 34.98 & 68.86 & 13.69 & 43.64 & 54.59 & 32.33 & 64.77 & 55.06 \\
MolMo 72B & 33.33 & 69.64 & 56.01 & 71.01 & 32.29 & 78.51 & 14.52 & 63.56 & 70.74 & 40.95 & 84.09 & 65.19 \\
Qwen2-VL 7B & 35.04 & 85.27 & 68.35 & 75.36 & 39.46 & 87.28 & 14.11 & 67.37 & 72.93 & 37.07 & 94.32 & 84.18 \\
Qwen2-VL 72B & 32.48 & 68.30 & 62.34 & 69.08 & 47.98 & 71.49 & 15.35 & 63.98 & 66.38 & 34.48 & 94.32 & 74.68 \\
Human & 40.00 & 82.00 & 96.00 & 68.00 & 40.00 & 92.00 & 14.00 & 84.00 & 88.00 & 67.00 & 100.00 & 33.33 \\ \bottomrule
\end{tabular}
}
\end{table}
\end{landscape}

\begin{landscape}
\begin{table}[]
\centering
\caption{Performance of all VLMs on every real-world dataset (Part 2).}
\label{tab:all_perf_2}
\resizebox{\textwidth}{!}{%
\begin{tabular}{@{}cccccccccccccc@{}}
\toprule
 & \textbf{Galaxy} & \textbf{Remote} & \textbf{OCRW} & \textbf{D\&T} & \textbf{VStar} & \textbf{MAME} & \textbf{Izem} & \textbf{ArtB} & \textbf{Grass} & \textbf{MMAD} & \textbf{Mile} & \textbf{PANDA} & \textbf{CAD} \\ \midrule
Random & 25.00 & 20.00 & 20.00 & 20.00 & 34.09 & 10.00 & 22.33 & 11.11 & 20.00 & 29.75 & 27.67 & 25.00 & 25.03 \\
Calude3 Haiku & 74.07 & 10.13 & 51.11 & 45.02 & 28.66 & 74.11 & 20.60 & 56.03 & 16.24 & 61.90 & 51.54 & 12.83 & 43.44 \\
Calude3.5 Sonnect & 59.26 & 32.60 & 67.11 & 71.43 & 46.34 & 83.04 & 22.32 & 61.64 & 17.09 & 68.40 & 62.56 & 16.81 & 81.45 \\
Gemini1.5 Pro & 81.48 & 37.00 & 75.11 & 51.95 & 65.24 & 83.04 & 19.31 & 69.40 & 32.48 & 70.13 & 59.47 & 18.58 & 67.42 \\
Gemini2.0 Flash & 51.85 & 48.02 & 75.11 & 76.62 & 71.95 & 86.61 & 22.32 & 59.91 & 29.91 & 71.00 & 66.52 & 30.09 & 83.26 \\
DeepSeek-VL2 27B & 81.48 & 54.19 & 63.56 & 56.28 & 67.07 & 75.45 & 30.47 & 64.66 & 21.37 & 73.59 & 60.35 & 34.51 & 75.57 \\
GPT-4o & 85.19 & 33.92 & 76.89 & 51.52 & 49.39 & 82.14 & 27.90 & 65.95 & 20.94 & 72.73 & 59.47 & 23.01 & 59.28 \\
GPT-4o mini & 77.78 & 13.66 & 60.89 & 44.59 & 51.22 & 75.45 & 24.46 & 59.05 & 16.67 & 71.43 & 56.39 & 15.49 & 46.61 \\
InternVL2 1B & 44.44 & 14.98 & 39.56 & 25.11 & 32.93 & 31.25 & 40.34 & 22.41 & 20.94 & 51.95 & 49.34 & 18.58 & 39.82 \\
InternVL2 2B & 66.67 & 37.44 & 57.78 & 35.06 & 52.44 & 55.36 & 18.88 & 53.02 & 16.24 & 54.98 & 63.88 & 22.12 & 40.27 \\
InternVL2 4B & 59.26 & 39.21 & 60.89 & 52.81 & 47.56 & 66.07 & 21.03 & 55.17 & 28.63 & 65.37 & 63.88 & 30.53 & 61.09 \\
InternVL2 8B & 70.37 & 34.80 & 68.89 & 35.50 & 64.02 & 66.52 & 18.88 & 56.90 & 32.05 & 66.67 & 64.76 & 18.14 & 60.63 \\
InternVL2 26B & 77.78 & 46.26 & 74.67 & 57.14 & 68.29 & 78.13 & 24.46 & 56.47 & 23.93 & 71.86 & 69.16 & 26.11 & 66.52 \\
InternVL2 40B & 74.07 & 50.66 & 77.78 & 58.01 & 75.00 & 82.14 & 16.31 & 65.09 & 34.19 & 74.89 & 72.69 & 23.01 & 76.02 \\
InternVL2 76B & 70.37 & 49.78 & 74.22 & 58.01 & 73.17 & 81.70 & 23.61 & 64.22 & 17.52 & 77.49 & 71.37 & 10.62 & 63.35 \\
DeepSeek-Janus 7B & 77.78 & 37.89 & 50.67 & 20.78 & 42.68 & 66.52 & 19.31 & 49.57 & 25.64 & 67.10 & 44.05 & 34.51 & 45.25 \\
Llama3.2 11B & 70.37 & 45.82 & 66.22 & 48.48 & 56.10 & 72.32 & 18.45 & 58.62 & 34.62 & 67.97 & 60.79 & 23.01 & 66.06 \\
Llama3.2 90B & 74.07 & 38.33 & 72.00 & 44.59 & 60.37 & 82.59 & 24.03 & 59.48 & 19.23 & 71.00 & 62.56 & 19.03 & 70.14 \\
Llava-HR 7B & 48.15 & 22.47 & 44.00 & 20.78 & 39.02 & 50.00 & 30.90 & 39.22 & 20.94 & 54.98 & 41.85 & 15.49 & 29.41 \\
Llava-Next 8B & 70.37 & 44.93 & 52.44 & 26.41 & 59.76 & 60.71 & 30.47 & 40.95 & 16.24 & 67.53 & 56.83 & 18.14 & 40.72 \\
Llava-Next 13B & 77.78 & 31.28 & 56.44 & 35.50 & 49.39 & 54.46 & 18.88 & 43.97 & 16.67 & 64.94 & 49.78 & 12.39 & 30.77 \\
Llava-Next 34B & 88.89 & 43.17 & 59.11 & 37.23 & 59.15 & 74.55 & 20.17 & 56.03 & 40.60 & 74.46 & 57.27 & 21.24 & 63.80 \\
Llava-Next 72B & 74.07 & 46.70 & 58.67 & 33.77 & 57.93 & 73.21 & 17.60 & 59.91 & 19.23 & 74.89 & 58.59 & 14.60 & 49.77 \\
Llava-OneVision 72B & 88.89 & 44.49 & 73.33 & 52.81 & 78.05 & 80.36 & 27.90 & 62.93 & 41.45 & 75.32 & 68.28 & 21.68 & 61.54 \\
Phi3.5 4B & 81.48 & 40.97 & 59.11 & 47.19 & 53.05 & 59.38 & 9.87 & 56.03 & 29.06 & 69.26 & 53.30 & 19.91 & 56.56 \\
MolMo 7B-D & 55.56 & 49.34 & 64.00 & 33.77 & 71.34 & 51.34 & 24.89 & 41.81 & 24.79 & 72.29 & 56.39 & 22.57 & 50.68 \\
MolMo 72B & 85.19 & 51.98 & 75.56 & 35.93 & 71.95 & 64.73 & 35.19 & 53.88 & 47.01 & 76.19 & 56.83 & 26.11 & 62.44 \\
Qwen2-VL 7B & 85.19 & 50.66 & 78.22 & 67.10 & 84.15 & 83.93 & 33.48 & 62.50 & 26.07 & 76.19 & 72.69 & 28.32 & 81.45 \\
Qwen2-VL 72B & 70.37 & 43.61 & 78.67 & 52.38 & 76.22 & 82.59 & 40.77 & 61.21 & 26.92 & 73.59 & 67.40 & 15.04 & 64.71 \\
Human & 54.00 & 87.00 & 68.00 & 93.00 & 100 & 76.00 & 20.00 & 57.00 & 43.00 & 75.00 & 86.00 & 46.00 & 93.00 \\ \bottomrule
\end{tabular}
}
\end{table}
\end{landscape}

\section{Prompts and Metrics}
\label{sec:metrics}
For ComplexGrid dataset, our prompt is ``The image is composed of multiple sub-images. The left upper corner is row 1 column 1. We also add the row and column numbers under each image. You need to identify the sub-image that best suits the caption: \{caption\}, returning the row and column id of the needle sub-image in this format: \textless row\textgreater ROW\textless /row\textgreater \textless col\textgreater COL\textless /col\textgreater, such as \textless row\textgreater 3\textless /row\textgreater \textless col\textgreater 2\textless /col\textgreater'' . We ask the model to answer with the HTML tag because we could use Beautifulsoup to parse the tag to get a clean prediction to avoid evaluation bias. For the real-world dataset, we also adopt the same idea as tag parse. Our prompt is ``{question}\\n Give an answer with this format: \textless ans\textgreater ANSWER\textless /ans\textgreater , no redundant words. For example: \textless ans\textgreater A\textless /ans\textgreater ''. We use exact math as our metrics during the evaluation. 
\section{Examples}
\label{sec:examples}
Table~\ref{tab:ArtBench} to \ref{tab:VisDiffBench} show examples from HRScene real-world datasets. We compress the images to display them in the paper.  

\begin{table}[h]
    \centering
    \scriptsize
    \caption{Example from HRScene -- ArtBench}
    \label{tab:ArtBench}
    \begin{tabular}{p{.49\columnwidth}p{.49\columnwidth}}
        \includegraphics[width=0.48\columnwidth]{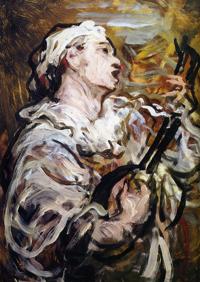}
        &
        \includegraphics[width=0.48\columnwidth]{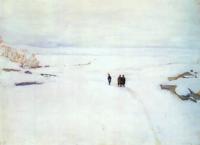} \\
        The painting in the picture belongs to which of the following categories?\newline
(A) Surrealism\newline
(B) Expressionism\newline
(C) Realism\newline
(D) Romanticism\newline
(E) Art Nouveau\newline
(F) Ukiyo E\newline
(G) Post Impressionism\newline
(H) Impressionism\newline
(I) Baroque
        &
        The painting in the picture belongs to which of the following categories?\newline
(A) Ukiyo E\newline
(B) Art Nouveau\newline
(C) Post Impressionism\newline
(D) Realism\newline
(E) Impressionism\newline
(F) Baroque\newline
(G) Romanticism\newline
(H) Expressionism\newline
(I) Surrealism \\
        \\
        Answer: H & Answer: E \\
    \end{tabular}
\end{table}

\begin{table}[h]
    \centering
    \scriptsize
    \caption{Example from HRScene -- Autonomous Driving}
    \label{tab:Autonomous_Driving}
    \begin{tabular}{p{.49\columnwidth}p{.49\columnwidth}}
        \includegraphics[width=0.48\columnwidth]{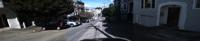}
        &
        \includegraphics[width=0.48\columnwidth]{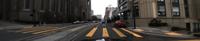} \\
        What is motion of the pedestrian wearing blue top on the left?\newline
(A) crossing the crosswalk\newline
(B) standing\newline
(C) jaywalking (illegally crossing not at pedestrian crossing)\newline
(D) walking on the sidewalk\newline
(E) The image does not feature the object
        &
        What is motion of the purple sedan on the right?\newline
(A) parked\newline
(B) moving\newline
(C) stopped\newline
(D) other\newline
(E) The image does not feature the object \\
        \\
        Answer: B & Answer: E \\
    \end{tabular}
\end{table}

\begin{table}[h]
    \centering
    \scriptsize
    \caption{Example from HRScene -- CAD}
    \label{tab:CAD}
    \begin{tabular}{p{.49\columnwidth}p{.49\columnwidth}}
        \includegraphics[width=0.48\columnwidth]{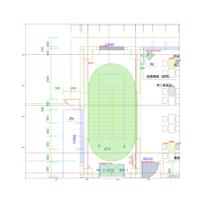}
        &
        \includegraphics[width=0.48\columnwidth]{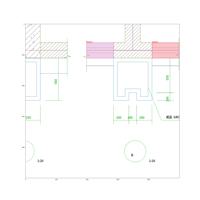} \\
        How many doors are there in the image?\newline
(A) 1\newline
(B) 0\newline
(C) 2\newline
(D) 3
        &
        What is the shape of the shadow at upper left corner of the image?\newline
(A) L-shape\newline
(B) Oval\newline
(C) Circle\newline
(D) Squre \\
        \\
        Answer: A & Answer: A \\
    \end{tabular}
\end{table}

\begin{table}[h]
    \centering
    \scriptsize
    \caption{Example from HRScene -- Diagram and Table}
    \label{tab:Diagram_and_Table}
    \begin{tabular}{p{.49\columnwidth}p{.49\columnwidth}}
        \includegraphics[width=0.48\columnwidth]{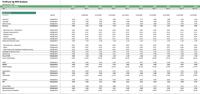}
        &
        \includegraphics[width=0.48\columnwidth]{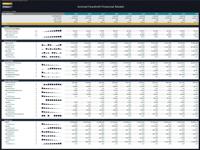} \\
        What's the data of Shipping Costs of 2028 Year 5 in the table Profit per kg NH3 Analysis?\newline
(A) -0.51\newline
(B) -0.52\newline
(C) -0.53\newline
(D) -0.54\newline
(E) This image doesn't feature the data.
        &
        What is the revenue of Pigs Feed in year 5 in the Revenue Sources table?\newline
(A) 4.548,625\newline
(B) 4.223.063\newline
(C) 3.710.817\newline
(D) 4.058.442\newline
(E) The image does not feature the number. \\
        \\
        Answer: D & Answer: D \\
    \end{tabular}
\end{table}

\begin{table}[h]
    \centering
    \scriptsize
    \caption{Example from HRScene -- DocStruct4M}
    \label{tab:DocStruct4M}
    \begin{tabular}{p{.49\columnwidth}p{.49\columnwidth}}
        \includegraphics[width=0.48\columnwidth]{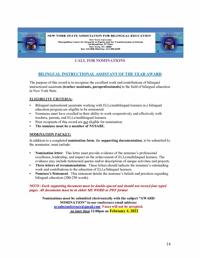}
        &
        \includegraphics[width=0.48\columnwidth]{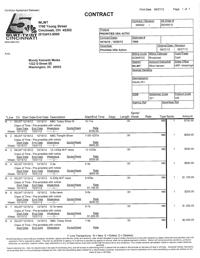} \\
        Read the following text: \textless doc\textgreater  CALL FOR NOMINATIONS \newline
 BILINGUAL INSTRUCTIONAL ASSISTANT OF THE YEAR AWARD \newline
\textcolor{red}{[omitted]}\newline
Which of the following options is correct?\newline
(A) the nominee’s outstanding 
\textcolor{red}{[omitted]}\newline
 14 \textless /doc\textgreater \newline
(B) the nominee’s outstanding 
\textcolor{red}{[omitted]}\newline
 14 \textless /doc\textgreater \newline
(C) the nominee’s outstanding 
\textcolor{red}{[omitted]}\newline
 14 \textless /doc\textgreater \newline
(D) the nominee’s outstanding 
\textcolor{red}{[omitted]}\newline
 14 \textless /doc\textgreater 
        &
        Which of the following sentences is present in the image?\newline
Which of the following options is correct?\newline
(A) \textless ocr\textgreater  CONTACTS \textless /ocr\textgreater \newline
(B) \textless ocr\textgreater  CONTACT \textless /ocr\textgreater \newline
(C) \textless ocr\textgreater  CONTRACT \textless /ocr\textgreater \newline
(D) \textless ocr\textgreater  CONVENANT \textless /ocr\textgreater  \\
        \\
        Answer: D & Answer: C \\
    \end{tabular}
\end{table}

\begin{table}[h]
    \centering
    \scriptsize
    \caption{Example from HRScene -- Galaxy}
    \label{tab:Galaxy}
    \begin{tabular}{p{.49\columnwidth}p{.49\columnwidth}}
        \includegraphics[width=0.48\columnwidth]{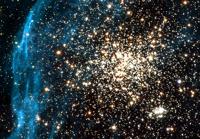}
        &
        \includegraphics[width=0.48\columnwidth]{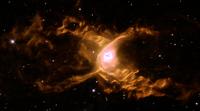} \\
        What type of celestial object is shown in the image? Please note that only clearly visible or distinguishable celestial bodies are counted.\newline
(A) Elliptical\newline
(B) star\newline
(C) Spiral\newline
(D) irregular
        &
        Does the galaxy have a distinct central core? Please note that only clearly visible or distinguishable celestial bodies are counted.\newline
(A) No\newline
(B) I don't know\newline
(C) Yes\newline
(D) two \\
        \\
        Answer: B & Answer: C \\
    \end{tabular}
\end{table}

\begin{table}[h]
    \centering
    \scriptsize
    \caption{Example from HRScene -- Grass}
    \label{tab:Grass}
    \begin{tabular}{p{.49\columnwidth}p{.49\columnwidth}}
        \includegraphics[width=0.48\columnwidth]{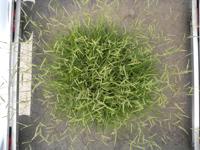}
        &
        \includegraphics[width=0.48\columnwidth]{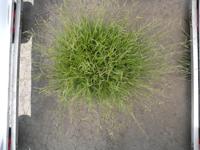} \\
        Based on the plant in the image, which growth stage does it belong to, and how many racemes does it have?\newline
(A) Reproductive stage, more than 200\newline
(B) Reproductive stage, 10-100 (include 100)\newline
(C) Reproductive stage, 0-10 (include 10)\newline
(D) Reproductive stage, 100-200 (include 200)\newline
(E) Vegetative stage, no racemes
        &
        Based on the plant in the image, which growth stage does it belong to, and how many racemes does it have?\newline
(A) Reproductive stage, 10-100 (include 100)\newline
(B) Reproductive stage, more than 200\newline
(C) Reproductive stage, 0-10 (include 10)\newline
(D) Vegetative stage, no racemes\newline
(E) Reproductive stage, 100-200 (include 200) \\
        \\
        Answer: A & Answer: B \\
    \end{tabular}
\end{table}

\begin{table}[h]
    \centering
    \scriptsize
    \caption{Example from HRScene -- HR-Bench}
    \label{tab:HR-Bench}
    \begin{tabular}{p{.49\columnwidth}p{.49\columnwidth}}
        \includegraphics[width=0.48\columnwidth]{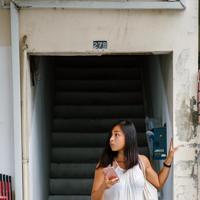}
        &
        \includegraphics[width=0.48\columnwidth]{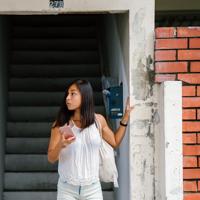} \\
        What is the number displayed above the entrance where the woman is standing?\newline
(A) 27E\newline
(B) 37B\newline
(C) 27D\newline
(D) 27B
        &
        What is the color of the mailbox?\newline
(A) Green\newline
(B) Black\newline
(C) Red\newline
(D) Blue \\
        \\
        Answer: D & Answer: D \\
    \end{tabular}
\end{table}

\begin{table}[h]
    \centering
    \scriptsize
    \caption{Example from HRScene -- HRIQ}
    \label{tab:HRIQ}
    \begin{tabular}{p{.49\columnwidth}p{.49\columnwidth}}
        \includegraphics[width=0.48\columnwidth]{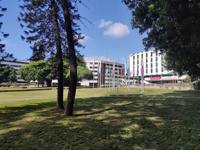}
        &
        \includegraphics[width=0.48\columnwidth]{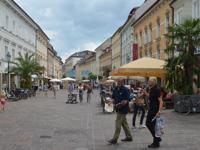} \\
        Assess the quality of a given image and predict a score that reflects the mean subjective human judgment of image quality. Some factors you may consider are distortions, such as Noise, Out-of-focus blur, Motion blur, Overexposure / Underexposure, Low contrast, Incorrect saturation, Sensor noise, and any combination of these distortions. Do not rely on metadata or external references - your judgment should be based purely on visual quality.\newline
(A) 1 bad\newline
(B) 2 poor\newline
(C) 3 fair\newline
(D) 4 good\newline
(E) 5 excellent
        &
        Assess the quality of a given image and predict a score that reflects the mean subjective human judgment of image quality. Some factors you may consider are distortions, such as Noise, Out-of-focus blur, Motion blur, Overexposure / Underexposure, Low contrast, Incorrect saturation, Sensor noise, and any combination of these distortions. Do not rely on metadata or external references - your judgment should be based purely on visual quality.\newline
(A) 1 bad\newline
(B) 2 poor\newline
(C) 3 fair\newline
(D) 4 good\newline
(E) 5 excellent \\
        \\
        Answer: D & Answer: D \\
    \end{tabular}
\end{table}

\begin{table}[h]
    \centering
    \scriptsize
    \caption{Example from HRScene -- InfographicVQA}
    \label{tab:InfographicVQA}
    \begin{tabular}{p{.49\columnwidth}p{.49\columnwidth}}
        \includegraphics[width=0.48\columnwidth]{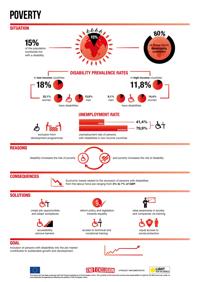}
        &
        \includegraphics[width=0.48\columnwidth]{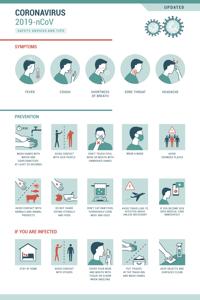} \\
        what percent of people live without disability around the world according to the data given? \newline
(A) '80, '80\%'\newline
(B) '79.9', '79.9\%'\newline
(C) '15', '15\%'\newline
(D) '85', '85\%'
        &
        Which of these animals are shown in the image? \newline
(A) 'cow, fish'\newline
(B) 'cow, human'\newline
(C) 'cat, cow'\newline
(D) 'plane, apple' \\
        \\
        Answer: D & Answer: A \\
    \end{tabular}
\end{table}

\begin{table}[h]
    \centering
    \scriptsize
    \caption{Example from HRScene -- Izembek}
    \label{tab:Izembek}
    \begin{tabular}{p{.49\columnwidth}p{.49\columnwidth}}
        \includegraphics[width=0.48\columnwidth]{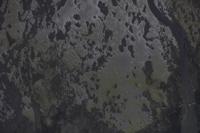}
        &
        \includegraphics[width=0.48\columnwidth]{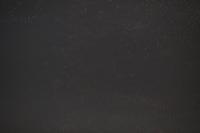} \\
        How many goose or other animals do you see in the image?\newline
(A) more than 400\newline
(B) 100-200\newline
(C) 200-300\newline
(D) 300-400
        &
        How many goose or other animals do you see in the image?\newline
(A) more than 400\newline
(B) 300-400\newline
(C) 200-300\newline
(D) 100-200 \\
        \\
        Answer: A & Answer: C \\
    \end{tabular}
\end{table}

\begin{table}[h]
    \centering
    \scriptsize
    \caption{Example from HRScene -- LungHist700}
    \label{tab:LungHist700}
    \begin{tabular}{p{.49\columnwidth}p{.49\columnwidth}}
        \includegraphics[width=0.48\columnwidth]{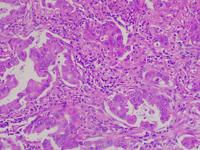}
        &
        \includegraphics[width=0.48\columnwidth]{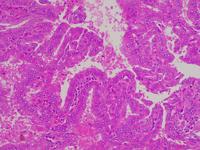} \\
        Given the following histopathological image of lung tissue, classify the malignancy (if any) into one of the seven categories:\newline
(A) Normal tissue\newline
(B) Adenocarcinoma (Well-differentiated)\newline
(C) Adenocarcinoma (Moderately differentiated)\newline
(D) Adenocarcinoma (Poorly differentiated)\newline
(E) Squamous cell carcinoma (Well-differentiated)\newline
(F) Squamous cell carcinoma (Moderately differentiated)\newline
(G) Squamous cell carcinoma (Poorly differentiated)
        &
        Given the following histopathological image of lung tissue, classify the malignancy (if any) into one of the seven categories:\newline
(A) Normal tissue\newline
(B) Adenocarcinoma (Well-differentiated)\newline
(C) Adenocarcinoma (Moderately differentiated)\newline
(D) Adenocarcinoma (Poorly differentiated)\newline
(E) Squamous cell carcinoma (Well-differentiated)\newline
(F) Squamous cell carcinoma (Moderately differentiated)\newline
(G) Squamous cell carcinoma (Poorly differentiated) \\
        \\
        Answer: B & Answer: B \\
    \end{tabular}
\end{table}

\begin{table}[h]
    \centering
    \scriptsize
    \caption{Example from HRScene -- MAME}
    \label{tab:MAME}
    \begin{tabular}{p{.49\columnwidth}p{.49\columnwidth}}
        \includegraphics[width=0.48\columnwidth]{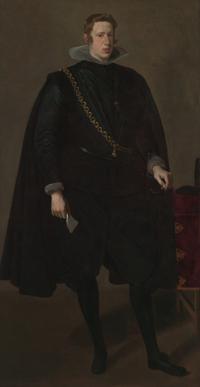}
        &
        \includegraphics[width=0.48\columnwidth]{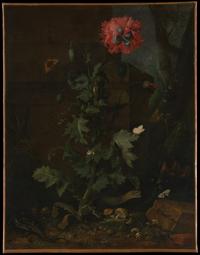} \\
        The artwork in the picture belongs to which of the following medium categories?\newline
(A) Hand-colored etching\newline
(B) Lithograph\newline
(C) Faience\newline
(D) Silk and metal thread\newline
(E) Graphite\newline
(F) Etching\newline
(G) Clay\newline
(H) Ivory\newline
(I) Woodcut\newline
(J) Oil on canvas
        &
        The artwork in the picture belongs to which of the following medium categories?\newline
(A) Lithograph\newline
(B) Oil on canvas\newline
(C) Ivory\newline
(D) Porcelain\newline
(E) Silver\newline
(F) Woodblock\newline
(G) Steel\newline
(H) Limestone\newline
(I) Marble\newline
(J) Iron \\
        \\
        Answer: J & Answer: B \\
    \end{tabular}
\end{table}

\begin{table}[h]
    \centering
    \scriptsize
    \caption{Example from HRScene -- MMAD}
    \label{tab:MMAD}
    \begin{tabular}{p{.49\columnwidth}p{.49\columnwidth}}
        \includegraphics[width=0.48\columnwidth]{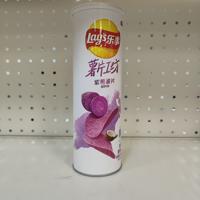}
        &
        \includegraphics[width=0.48\columnwidth]{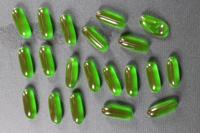} \\
        There is a defect in the object. Where is the defect?\newline
(A) On the top of the can\newline
(B) On the bottom of the can\newline
(C) Around the center region of the can, on the image of the potato chip\newline
(D) On the side of the can
        &
        There is a defect in the object. What is the appearance of the defect?\newline
(A) The defective capsule has a distinct non-conforming orange color.\newline
(B) The defective capsule has a shiny, translucent quality.\newline
(C) The defective capsule has a misshapen appearance.\newline
(D) The defective capsule has visible bubbles. \\
        \\
        Answer: C & Answer: A \\
    \end{tabular}
\end{table}

\begin{table}[h]
    \centering
    \scriptsize
    \caption{Example from HRScene -- MuirBench}
    \label{tab:MuirBench}
    \begin{tabular}{p{.49\columnwidth}p{.49\columnwidth}}
        \includegraphics[width=0.48\columnwidth]{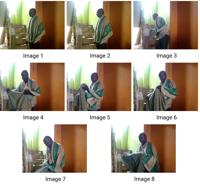}
        &
        \includegraphics[width=0.48\columnwidth]{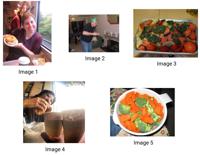} \\
        What type of clothing was the man primarily seen wearing? \textless image1 \textgreater \textless image2 \textgreater \textless image3 \textgreater \textless image4 \textgreater \textless image5 \textgreater \textless image6 \textgreater \textless image7 \textgreater \textless image8\textgreater \newline
(A) None of the choices provided\newline
(B) Green and white jacket\newline
(C) Robe and shawl\newline
(D) Sweater
        &
        \textless image1\textgreater  Which of the following images shares the same scene with the given image but contains the object dining table?\newline
(A) \textless image2\textgreater \newline
(B) \textless image3\textgreater \newline
(C) \textless image4\textgreater \newline
(D) None of the choices provided\newline
(E) \textless image5\textgreater  \\
        \\
        Answer: C & Answer: C \\
    \end{tabular}
\end{table}

\begin{table}[h]
    \centering
    \scriptsize
    \caption{Example from HRScene -- NovaChart}
    \label{tab:NovaChart}
    \begin{tabular}{p{.49\columnwidth}p{.49\columnwidth}}
        \includegraphics[width=0.48\columnwidth]{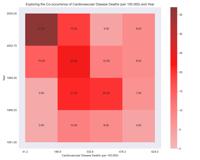}
        &
        \includegraphics[width=0.48\columnwidth]{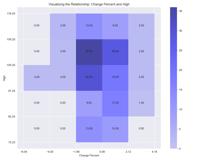} \\
        Can you discern the type of chart used in this visualization? From the provided alternatives, please select the correct choice for the question above:  \newline
(A) bivariate histogram\newline
(B) single-class scatter plot\newline
(C) radar chart\newline
(D) pie chart\newline
(E) univariate histogram
        &
        Can you provide the histogram value for the bin corresponding to the range x=[-4.0, -1.96) and y=[73.235, 82.2385)?\newline
(A) 13\newline
(B) 9\newline
(C) 2\newline
(D) 3\newline
(E) 0 \\
        \\
        Answer: A & Answer: E \\
    \end{tabular}
\end{table}

\begin{CJK}{UTF8}{min}

\begin{table}[h]
    \centering
    \scriptsize
    \caption{Example from HRScene -- OCR in the Wild}
    \label{tab:OCR_in_the_Wild}
    \begin{tabular}{p{.49\columnwidth}p{.49\columnwidth}}
        \includegraphics[width=0.48\columnwidth]{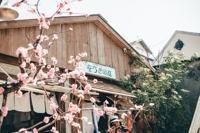}
        &
        \includegraphics[width=0.48\columnwidth]{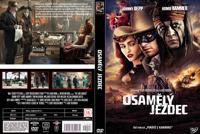} \\
        What is the content on the plaque in the center of the picture?\newline
(A) 安らぎの梃\newline
(B) 安らぎの廷\newline
(C) 安らぎの挺\newline
(D) 安らぎの庭\newline
(E) This image doesn't feature the content.
        &
        How long is this film in the picture?\newline
(A) 5.1\newline
(B) 2013\newline
(C) 148 min.\newline
(D) 143 min.\newline
(E) The image does not feature the content. \\
        \\
        Answer: D & Answer: D \\
    \end{tabular}
\end{table}

\end{CJK}

\begin{table}[h]
    \centering
    \scriptsize
    \caption{Example from HRScene -- PANDA}
    \label{tab:PANDA}
    \begin{tabular}{p{.49\columnwidth}p{.49\columnwidth}}
        \includegraphics[width=0.48\columnwidth]{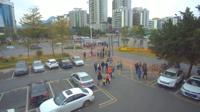}
        &
        \includegraphics[width=0.48\columnwidth]{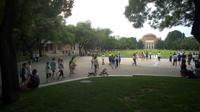} \\
        How many riding person(s) are in the image?\newline
(A) 35\newline
(B) 27\newline
(C) 23\newline
(D) 44
        &
        How many riding person(s) are in the image?\newline
(A) 11\newline
(B) 12\newline
(C) 21\newline
(D) 15 \\
        \\
        Answer: C & Answer: B \\
    \end{tabular}
\end{table}

\begin{table}[h]
    \centering
    \scriptsize
    \caption{Example from HRScene -- Remote Sensing}
    \label{tab:Remote_Sensing}
    \begin{tabular}{p{.49\columnwidth}p{.49\columnwidth}}
        \includegraphics[width=0.48\columnwidth]{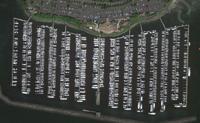}
        &
        \includegraphics[width=0.48\columnwidth]{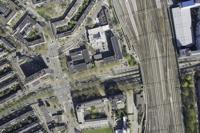} \\
        What color is the second ship from top to bottom on the far right side of the picture?\newline
(A) White\newline
(B) Red\newline
(C) Green\newline
(D) Yellow\newline
(E) This image doesn't feature the color.
        &
        How many red cars are there in the parking lot in the middle of the bottom of the picture?\newline
(A) 1\newline
(B) 2\newline
(C) 3\newline
(D) 4\newline
(E) This image doesn't feature the count. \\
        \\
        Answer: A & Answer: D \\
    \end{tabular}
\end{table}

\begin{table}[h]
    \centering
    \scriptsize
    \caption{Example from HRScene -- VQA-RAD}
    \label{tab:VQA-RAD}
    \begin{tabular}{p{.49\columnwidth}p{.49\columnwidth}}
        \includegraphics[width=0.48\columnwidth]{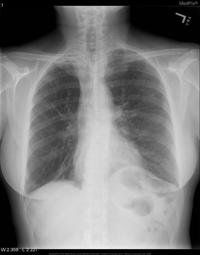}
        &
        \includegraphics[width=0.48\columnwidth]{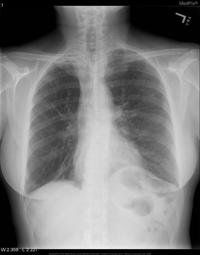} \\
        Is the trachea midline?\newline
(A) Yes\newline
(B) No\newline
(C) Not specified
        &
        Is there evidence of an aortic aneurysm?\newline
(A) Yes\newline
(B) No\newline
(C) Not specified \\
        \\
        Answer: A & Answer: B \\
    \end{tabular}
\end{table}

\begin{table}[h]
    \centering
    \scriptsize
    \caption{Example from HRScene -- VStar Bench}
    \label{tab:VStar_Bench}
    \begin{tabular}{p{.49\columnwidth}p{.49\columnwidth}}
        \includegraphics[width=0.48\columnwidth]{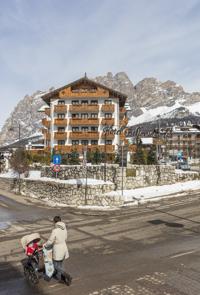}
        &
        \includegraphics[width=0.48\columnwidth]{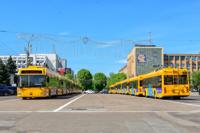} \\
        What is the color of the car?\newline
(A) The color of the car is silver.\newline
(B) The color of the car is black.\newline
(C) The color of the car is red.\newline
(D) The color of the car is blue.
        &
        Is the flag blue and yellow or red and yellow?\newline
(A) The color of the flag is red and yellow.\newline
(B) The color of the flag is blue and yellow. \\
        \\
        Answer: A & Answer: B \\
    \end{tabular}
\end{table}

\begin{table}[h]
    \centering
    \scriptsize
    \caption{Example from HRScene -- Video Monitoring}
    \label{tab:Video_Monitoring}
    \begin{tabular}{p{.49\columnwidth}p{.49\columnwidth}}
        \includegraphics[width=0.48\columnwidth]{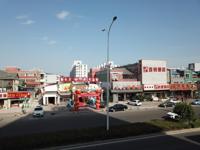}
        &
        \includegraphics[width=0.48\columnwidth]{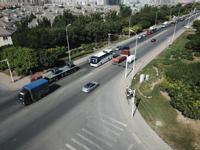} \\
        What is the number of people in the image?(If a human maintains standing pose or walking, please classify it as pedestrian, otherwise, it is classified as a people.)\newline
(A) 97\newline
(B) 88\newline
(C) 52\newline
(D) 100\newline
(E) The image does not feature the people
        &
        What is the number of tricycles in the image?\newline
(A) 51\newline
(B) 97\newline
(C) 55\newline
(D) 74\newline
(E) The image does not feature the tricycles \\
        \\
        Answer: E & Answer: E \\
    \end{tabular}
\end{table}

\begin{table}[h]
    \centering
    \scriptsize
    \caption{Example from HRScene -- VisDiffBench}
    \label{tab:VisDiffBench}
    \begin{tabular}{p{.49\columnwidth}p{.49\columnwidth}}
        \includegraphics[width=0.48\columnwidth]{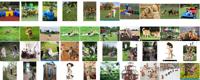}
        &
        \includegraphics[width=0.48\columnwidth]{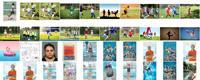} \\
        What is the difference between the first two rows of images and the last two rows? \newline
(A) Animal species (Dogs vs Cats)\newline
(B) Animal species (Cows vs Cats)\newline
(C) Background Colors (Green vs Blue)\newline
(D) Number of Objects (2 vs 3)
        &
        What is the difference between the first two rows of images and the last two rows?\newline
(A) Activity (Basketball vs Swimming)\newline
(B) Number of animals (1 vs 2)\newline
(C) Animal species (Cows vs Cats)\newline
(D) Activity (Soccer vs Swimming) \\
        \\
        Answer: A & Answer: D \\
    \end{tabular}
\end{table}

\end{document}